\documentclass[11pt]{article}
\pdfoutput=1

\usepackage{microtype}
\usepackage{graphicx}
\usepackage{subfigure}
\usepackage{booktabs} 
\usepackage{amssymb}
\usepackage{bm}
\usepackage{bbm}
\usepackage{amsmath}  
\usepackage{amsthm}
\usepackage{xcolor}
\usepackage{textcomp}
\usepackage{multirow}
\usepackage{mathtools}
\usepackage[margin=1in]{geometry}
\usepackage{authblk}
\usepackage[numbers]{natbib}
\usepackage{xspace}
\usepackage{comment}
\usepackage{thmtools, thm-restate}  
\usepackage{verbatim}
\usepackage{subcaption}

\usepackage{tikz}
\usetikzlibrary{arrows}
\usetikzlibrary{positioning}

\usepackage{array}
\usepackage{listings}
\usepackage{xcolor}
\usepackage{float}
\usepackage{wrapfig}
\usepackage{subfigure}

\definecolor{codegreen}{rgb}{0,0.6,0}
\definecolor{codegray}{rgb}{0.5,0.5,0.5}
\definecolor{codepurple}{rgb}{0.58,0,0.82}
\definecolor{backcolour}{rgb}{0.95,0.95,0.92}

\lstdefinestyle{mystyle}{
    backgroundcolor=\color{backcolour},   
    commentstyle=\color{codegreen},
    keywordstyle=\color{magenta},
    numberstyle=\tiny\color{codegray},
    stringstyle=\color{codepurple},
    basicstyle=\ttfamily\footnotesize,
    breakatwhitespace=false,         
    breaklines=true,                 
    captionpos=b,                    
    keepspaces=true,                 
    numbers=left,                    
    numbersep=5pt,                  
    showspaces=false,                
    showstringspaces=false,
    showtabs=false,                  
    tabsize=2
}

\usepackage{hyperref}
\usepackage{caption}

\usepackage{bm}
\usepackage{enumitem}

\newcommand{\R}{\mathbb{R}}

\newcommand{\jjpar}[1]{\left( #1 \right)}

\newcommand{\softmax}[1]{\text{softmax}\jjpar{#1}}
\newcommand{\SDP}[1]{\text{SDP}\jjpar{#1}}
\newcommand{\LSE}[1]{\text{LSE}\jjpar{#1}}

\title{Hydragen: High-Throughput LLM Inference with Shared Prefixes}
  \author[$\dagger$]{Jordan Juravsky\thanks{Equal Contribution.}}
  \author[$\ddagger$]{Bradley Brown$^*$}
  \author[$\S$]{Ryan Ehrlich$^*$}
  \author[$\dagger$]{Daniel Y. Fu}
  \author[$\dagger$]{Christopher R{\'e}}
  \author[$\dagger$]{Azalia Mirhoseini}
  \affil[$\dagger$]{Department of Computer Science, Stanford University}
  \affil[$\ddagger$]{University of Oxford}
  \affil[$\S$]{University of Waterloo\vspace{4pt}}
  \affil[ ]{{\texttt{jbj@stanford.edu}, \texttt{bradley.brown@cs.ox.ac.uk}, \texttt{rehrlich@uwaterloo.ca}, \texttt{danfu@stanford.edu}, \texttt{chrismre@stanford.edu}}, \texttt{azalia@stanford.edu}}

  \date{}
\begin{document}

\maketitle

\begin{abstract}
Transformer-based large language models (LLMs) are now deployed to hundreds of millions of users. LLM inference is commonly performed on batches of sequences that share a prefix, such as few-shot examples or a chatbot system prompt. Decoding in this large-batch setting can be bottlenecked by the attention operation, which reads large key-value (KV) caches from memory and computes inefficient matrix-vector products for every sequence in the batch. In this work, we introduce Hydragen, a hardware-aware exact implementation of attention with shared prefixes. Hydragen computes attention over the shared prefix and unique suffixes separately. This decomposition enables efficient prefix attention by batching queries together across sequences, reducing redundant memory reads and enabling the use of hardware-friendly matrix multiplications. Our method can improve end-to-end CodeLlama-13b throughput by up to 32x against competitive baselines, with speedup growing with the batch size and shared prefix length. Hydragen also enables the use of very long shared contexts: with a large batch size, increasing the prefix length from 1K to 16K tokens decreases Hydragen throughput by less than 15\%, while the throughput of baselines drops by over 90\%. Hydragen generalizes beyond simple prefix-suffix decomposition and can be applied to tree-based prompt sharing patterns, allowing us to further reduce inference time on competitive programming problems by 55\%. Our code is available at \url{https://github.com/jordan-benjamin/hydragen}.
\end{abstract}

\section{Introduction}
\begin{figure*}
    \centering
    \includegraphics[width=\textwidth]{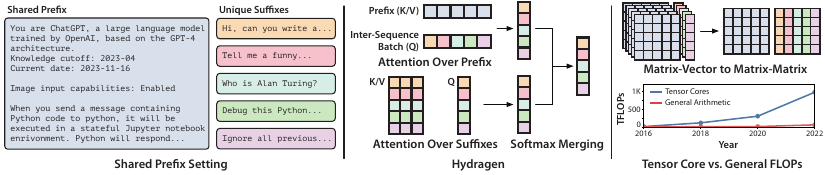}
    \caption{
    Left: An example LLM inference scenario where a chatbot model processes many sequences that share a large shared prefix (the system prompt). Middle: An overview of Hydragen, where overall attention is decomposed into attention over the shared prefix (batched across all queries in a batch) and attention over the remaining suffixes (independent across sequences, as is normally done). Top Right: Hydragen's attention decomposition allows many matrix vector products to be replaced with fewer matrix-matrix products. Bottom Right: Using matrix-matrix products is particularly important as GPUs dedicate an increasingly large ratio of their total FLOPS to tensor cores that are specialized in matrix multiplication.
    }
    \label{fig:banner}
\end{figure*}

Text generation on a batch of sequences is a common setting for LLM inference. In many real-world use cases, sequences in a batch share a common prefix. Examples include a chatbot serving many users with shared system instructions (Figure~\ref{fig:banner} left), an assistant model using a few-shot prompt for solving domain-specific tasks \cite{brown2020gpt3}, and competitive programming systems that sample many candidate solutions for a single problem \cite{Li_2022alphacode}. As transformer-based LLMs  \cite{vaswani2023attention} are deployed at increasingly large scales \cite{OpenAIChatGPT2023}, improving their efficiency with shared prefixes can have a significant impact. In this work, we use a hardware-aware perspective to analyze and optimize this inference setting.

Shared prefixes create overlaps in the attention keys and values across sequences, presenting an opportunity for specialized optimization. Existing work \cite{kwon2023vllm} identifies that naive KV caching leads to redundant storage of the prefix’s keys and values, and addresses this redundancy with a paged memory management strategy. While this optimization can significantly reduce GPU memory consumption, it does little to affect the speed of computing attention, which can often bottleneck end-to-end throughput with large batches. Since each sequence in the batch has a distinct (albeit overlapping) KV cache but only a single attention query when decoding, existing attention implementations like FlashAttention \cite{dao2022flashattention, dao2023flashattention2} and PagedAttention \cite{kwon2023vllm} compute attention by performing many independent matrix-vector products. This approach is memory-bound when the KV cache is large, and moreover does not use hardware-friendly matrix multiplications. Both of these characteristics lead to poor performance on modern GPUs. Across successive hardware generations, GPU computational capability has improved at a significantly faster rate than memory bandwidth. 
Additionally, an increasingly large fraction of total GPU floating-point operations (FLOPs) are only available when using tensor cores, a specialized hardware feature that is dedicated to performing matrix-matrix products and not matrix-vector products (Figure~\ref{fig:banner} bottom right).

In this paper, we demonstrate that shared prefixes enable more than just memory savings, and can additionally be used to improve decoding throughput. We identify that FlashAttention and PagedAttention redundantly read the prefix’s keys and values from GPU memory when computing attention, regardless of whether the prefix is redundantly stored. In order to eliminate these redundant reads, we present Hydragen, an exact implementation of attention that is specialized for shared prefixes (Figure~\ref{fig:banner} middle). Hydragen decomposes full-sequence attention into separate attention computations over the prefix and suffixes. These sub-computations can be cheaply combined to recover the overall attention result (Section~\ref{sec:decomposition}). With attention decomposition, Hydragen is able to efficiently compute attention over the prefix by batching together attention queries across sequences (Section~\ref{sec:prefix-attention}). This inter-sequence batching replaces many matrix-vector products with fewer matrix-matrix products (Figure~\ref{fig:banner} top right), reducing redundant reads of the prefix and enabling the use of tensor cores.

Experimentally, we find that Hydragen can significantly improve LLM throughput in large-batch settings with shared prefixes. In end-to-end benchmarks, Hydragen increases the throughput of CodeLlama-13b~\cite{rozière2023code} by up to 32x over vLLM~\cite{kwon2023vllm}, a high-performance inference framework that avoids redundant prefix storage but not redundant prefix reads. The attention operation in isolation can be accelerated by over 16x using Hydragen when compared to a state-of-the-art FlashAttention baseline, with benefits increasing as the batch size and shared prefix length grow. We also demonstrate that Hydragen’s efficient processing of shared prefixes can influence algorithmic decisions on how to use LLMs most effectively. With a large batch size, Hydragen allows the shared prefix to grow from 1K tokens to 16K tokens with less than a 15\% throughput penalty whereas vLLM throughput decreases by over 90\%. On long document question answering tasks, we show that Hydragen can process 256 questions in less time than it takes a FlashAttention baseline to process 64 (Section~\ref{sec:long-doc}). Finally, we demonstrate that Hydragen’s attention decomposition and batching apply to more general patterns of prompt sharing than a single prefix-suffix split. When solving APPS competitive programming problems \cite{hendrycksapps2021}, where two levels of prompt sharing occur, we apply Hydragen hierarchically to maximize sharing and reduce evaluation time by an additional 55\% over a single-level of prompt sharing (Section~\ref{sec:exps-hierarchy}).

\section{Background}

\subsection{Hardware Efficiency Considerations}

\textbf{GPU Performance Bottlenecks:} GPUs possess a limited number of processors for performing computation and a limited amount of bandwidth for transferring data between processors and memory. When a program running on a GPU is bottlenecked waiting for compute units to finish processing, it can be classified as compute-bound. Alternatively, memory-bound programs are bottlenecked accessing GPU memory. To summarize a program's use of hardware resources, we can calculate its arithmetic intensity, defined as the total number of arithmetic operations performed divided by the total number of bytes transferred. Higher arithmetic intensities imply a greater use of computational resources relative to memory bandwidth.

\textbf{Batching:} Batching is a common optimization that can increase an operation's arithmetic intensity and reduce memory bottlenecks. Consider the example of computing matrix-vector products. To compute one product, each element of the input matrix is read from memory but is used in only a single multiply-accumulate. Therefore, the arithmetic intensity of the operation is low, and is memory-bound on GPUs. However, if many matrix-vector products need to be computed using the same matrix, we can batch the operations together into a single matrix-matrix product. In the batched operation, the cost of reading the input matrix is amortized over the batch of vectors. Each element of the input matrix is now used for many multiply-accumulates, increasing the arithmetic intensity of the overall operation and improving hardware utilization. 

\textbf{Tensor Cores:} Modern GPUs (and other AI accelerators) are designed with specialized units for efficiently computing matrix multiplications. Effectively using these resources can be crucial for achieving good overall performance; on GPUs, tensor cores dedicated to matrix multiplications can compute over 10x more floating-point operations per second (FLOPS) than the rest of the GPU. This further motivates batching matrix-vector products into matrix-matrix products.

\subsection{Attention and LLM Inference}

The focus of this work is optimizing attention in transformer-based LLMs. Scaled-dot-product (SDP) attention operates on a sequence of queries $Q \in \R^{N_q \times d}$, keys $K \in \R^{N_{kv} \times d}$, and values $V \in \R^{N_{kv} \times d}$, producing an output $O \in \R^{N_q \times d}$ defined as:

\begin{align}
    O = \text{SDP}(Q, K, V) &= \softmax{ \frac{QK^T} {\sqrt{d}} }V
    \label{eq:sdp}
\end{align}

We are particularly interested in the performance characteristics of attention during LLM text generation. Generation begins with a prefill stage that processes the starting sequence of tokens that the LLM will complete. The prefill phase encodes the entire prompt in parallel using a single transformer forward pass. Therefore, when computing attention we have $N_q = N_{kv} \gg 1$ and as a result the multiplications in Equation~\ref{eq:sdp} involving $K^T$ and $V$ are hardware-friendly matrix multiplications. After the prefill stage, completion tokens are iteratively decoded from the model, with one decoding step producing one new token and requiring one  forward pass. Decoding is accelerated by the use of a KV cache, which stores the attention keys and values of all previous tokens in the sequence. The KV cache avoids the need for reprocessing the entire sequence during every decoding step, and instead only the most recent token is passed through the model. However, this leads to an attention computation where $N_q = 1$ while $N_{kv} \gg 1$, making the multiplications with $K^T$ and $V$ matrix-vector products. Attention during decoding is therefore memory-bound and does not use tensor cores.

\subsection{Batched Inference}
\label{sec:batched-inference}

LLM inference throughput can be increased by generating text for a batch of sequences in parallel. With batched decoding, each forward pass of the model processes the most recent token from many sequences instead of only one. This batching increases the arithmetic intensity of transformer components such as the multilayer perceptron (MLP) blocks and allows these modules to use hardware-friendly matrix multiplications. However, batched text generation does not increase the intensity of attention, since every sequence has a distinct key and value matrix. Therefore, while other model components are able to use tensor cores during batched decoding, attention must be computed using many independent matrix-vector products. With large batch sizes or long sequence lengths, computing attention becomes increasingly expensive relative to rest of the transformer, decreasing throughput. Additionally, the storage footprint of the KV cache in GPU memory can exceed that of the model parameters when the batch size is large, imposing constraints on the maximum number of sequences that can be simultaneously processed.

\subsection{Shared Prefixes}
\label{sec:shared-prefixes}

In this paper, we investigate improving the throughput of batched text generation when the sequences in the batch share a common prefix. This scenario lends itself to specialized optimizations because shared prefixes lead to overlaps across sequences' key and value matrices. The causal attention mask in LLMs results in each token's activations being influenced only by previous tokens in the sequence. Therefore, if multiple sequences share a common prefix, the keys and values corresponding to the prefix tokens will be identical across sequences.

The key-value overlap introduced by shared prefixes presents two distinct directions for improving the inference process described in Section~\ref{sec:batched-inference}. Firstly, naive batched inference stores the KV cache separately for every sequence, leading to redundant storage of the prefix key and value vectors. Existing work has identified this redundancy and proposed an elegant virtual memory system to eliminate duplicate storage \cite{kwon2023vllm}.

In this work, we identify an additional opportunity to optimize the attention operation itself. When GPU kernels compute attention for each sequence in the batch using independent matrix-vector products, the prefix keys and values are repeatedly read from GPU memory, regardless of whether they are stored redundantly or not. We now propose an alternative approach to computing attention, which can simultaneously eliminate these redundant reads and enable the use of tensor cores.

\section{Hydragen: Efficient Attention with Shared Prefixes}

We introduce Hydragen, an exact implementation of attention that is optimized for shared prefixes. Hydragen is a combination of two techniques:

\begin{enumerate}
    \item \textbf{Attention Decomposition:} We split full-sequence attention into separate attention computations over the shared prefix and unique suffixes that can be cheaply combined to recover the full attention result.
    \item \textbf{Inter-Sequence Batching:} We efficiently compute attention over the prefix by batching together attention queries across sequences.
\end{enumerate}

Attention decomposition allows us to isolate overlapping portions of the batch's key and value matrices, while inter-sequence batching exploits this overlap by replacing many matrix-vector products with a single matrix-matrix product. Pseudocode implementing Hydragen attention is provided in Appendix~\ref{app:code}. 

\subsection{Decomposing Attention Across Subsequences}
\label{sec:decomposition}

As discussed in Section~\ref{sec:shared-prefixes}, sequences that share a common prefix have partially overlapping keys and values when computing attention. Our goal is to separate this computation with partial overlap into two separate operations: attention over the shared prefix, where there is total key-value overlap, and attention over unique suffixes, where there is no overlap.

Consider the general case where our keys $K$ and values $V$ are partitioned across $N_{kv}$ (the sequence/row dimension) into:

\begin{align}
    K &= K_1 || K_2 \\
    V &= V_1 || V_2
\end{align}

with $||$ denoting concatenation. We wish to avoid directly computing our desired quantity $\SDP{Q, K, V}$, and instead calculate this value using the results of the sub-computations $\SDP{Q, K_1, V_1}$ and $\SDP{Q, K_2, V_2}$.

The challenge in partitioning attention is with the softmax operation, since the softmax denominator is calculated by summing over all exponentiated attention scores in the sequence. In order to combine our sub-computations, we use a denominator rescaling trick inspired by FlashAttention's blocked softmax computation \cite{dao2022flashattention}. When computing $\SDP{Q, K_1, V_1}$ and $\SDP{Q, K_2, V_2}$, we additionally compute and store the log-sum-exp (LSE($Q, K$) $\in \R^{N_q}$) of the attention scores (equivalently, the log of the softmax denominator):

\begin{align}
    \LSE{Q, K} &= \log\jjpar{\text{sum}\jjpar{\exp\jjpar{\frac{QK^T}{\sqrt{d}}}, \text{dim}=1}}
\end{align}

Given the two partitioned attention outputs and their LSEs, we can calculate our final result $\SDP{Q, K, V}$ by computing the full-sequence softmax denominator and rescaling the attention outputs accordingly:

\begin{align}
    \text{SDP}(Q, K, V) &= \frac{\SDP{Q, K_1, V_1}
 e^{\LSE{Q, K_1}} + \SDP{Q, K_2, V_2}
 e^{\LSE{Q, K_2}}  }{e^{\LSE{Q, K_1}} + e^{\LSE{Q, K_2}}}
 \label{eq:decomposition}
\end{align}

We prove this formula in Appendix~\ref{app:decomposition}.

\subsection{Inter-Sequence Batched Prefix Attention}
\label{sec:prefix-attention}

With attention decomposition, we are able to compute attention over the prefix as a standalone operation for every sequence. While this decomposition does not improve performance on its own (in fact, it introduces additional work in order to combine sub-computation outputs), it can allow us to compute prefix attention much more efficiently over a batch of sequences. 

Queries do not affect each other when computing attention, therefore if two sets of queries attend over identical keys and values, they can be merged into a single attention operation with a larger number of queries. With attention decomposition, this case now applies to each sequence's attention over the shared prefix. Since the prefix's keys and values across sequences are identical, we can batch each sequence's query vector together into one attention operation over a single sequence. Importantly, this batching significantly raises $N_q$ and the arithmetic intensity of prefix attention, replacing many separate matrix-vector products with a single matrix-matrix product. By replacing multiple independent attention computations over the prefix with a single batched operation, we can reduce the number of times that the prefix KV cache is read from GPU memory. Additionally, we can now use tensor cores during prefix attention and significantly improve hardware utilization. 

Note that we are unable to apply inter-sequence batching when computing attention over suffixes, since the keys and values in each sequence's suffix are not identical. Suffix attention is therefore computed normally, with a single query per sequence.

\subsection{Hierarchical Sharing}
\label{sec:hierarchy}

\begin{figure}
    \centering
    \includegraphics[width=0.5\textwidth]{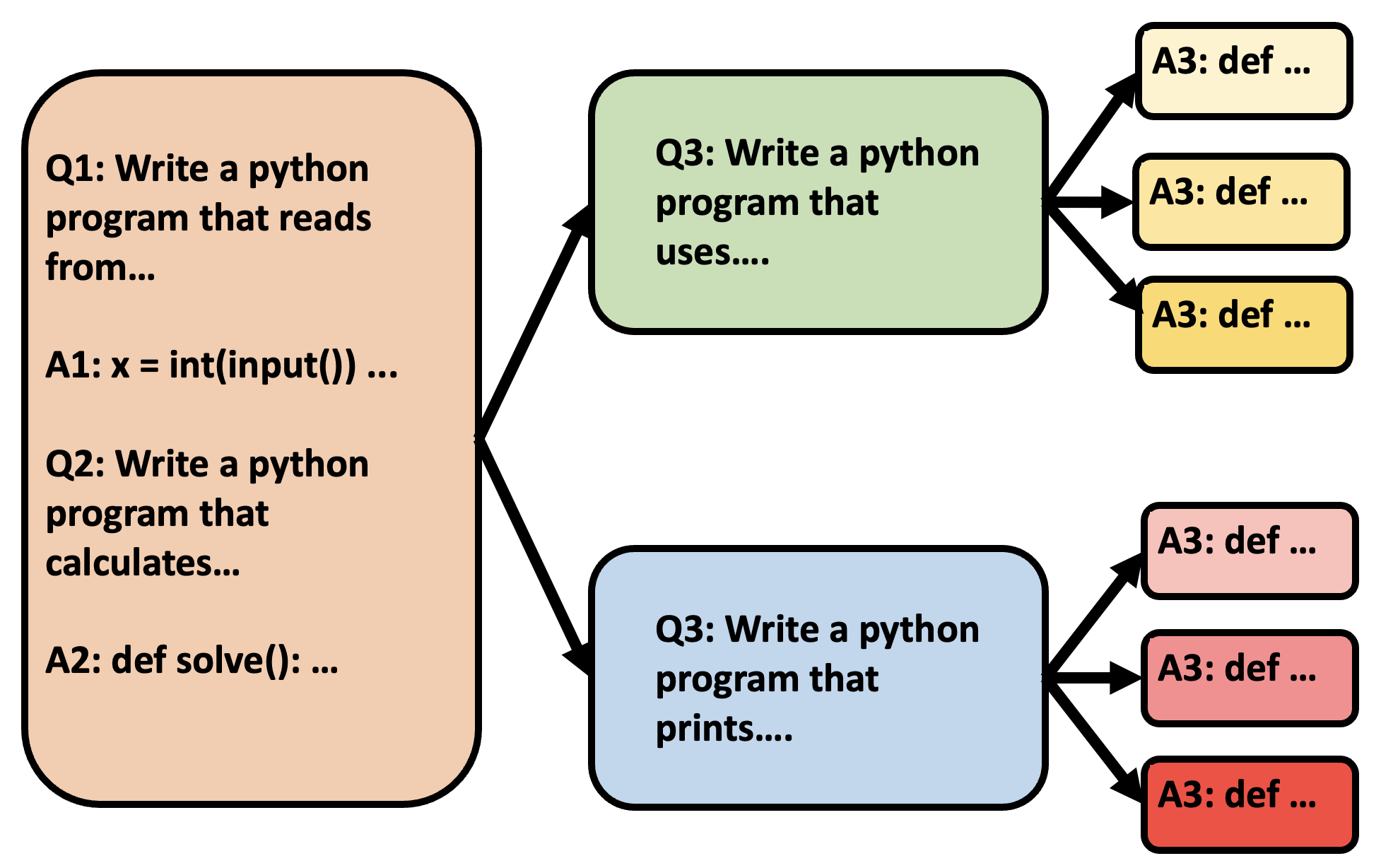}
    \caption{An example of a batch of sequences with a hierarchical sharing pattern. This diagram depicts the setting of Section~\ref{sec:exps-hierarchy}, which solves competitive programming problems using a few-shot prompt and by sampling many candidate solutions per problem. The few-shot prompt (orange) is globally shared across all sequences in the batch. However, the descriptions of each problem (green and blue) are only shared across the candidate solutions corresponding to that problem.}
    \label{fig:tree}
    \vspace{-5mm}
\end{figure}

So far, we have focused on the setting where all sequences in the batch share a common starting subsequence followed by suffixes that are distinct from one another. However, this excludes other forms of sharing that appear in important use cases. Sequences in the batch may not all start with a global prefix, and instead the batch may be divided into groups of overlapping sequences. Additionally, sharing may be more fine-grained than a simple prefix-suffix decomposition, with the overlap between sequences forming a tree structure where each node contains a token sequence that is shared by all descendants (see Figure~\ref{fig:tree} for an example). These forms of sharing are increasingly relevant as LLMs are applied in more complicated inference/search algorithms~\cite{yao2023tree, besta2023graph, ning2023skeletonofthought}. 

Hydragen naturally generalizes to these richer forms of sharing as well. To apply Hydragen to a tree of sequences, we replace attention decomposition over the prefix and suffix with attention decomposition at every vertex in the tree. We can then use inter-sequence batching across levels of the tree, so that the keys and values associated with one node in the tree are shared across the queries of all descendant nodes. 

\subsection{Estimating Throughput Improvements with Hydragen}

Hydragen significantly improves the efficiency of attention with shared prefixes relative to approaches that compute attention independently for every sequence (Section~\ref{sec:micro}). However, translating this targeted efficiency into end-to-end throughput improvements depends strongly on the details of the inference setting being considered. In order for Hydragen to meaningfully improve decoding speed in a particular setting, attention must be a major contributor to decoding time. For example, with small batch sizes or short sequence lengths, decoding speed is often bottlenecked not by attention, but by reading the parameters of the model from GPU memory. The benefits of Hydragen in this scenario will therefore be minimal. Similarly, given a fixed batch size and sequence length, we expect Hydragen to improve throughput more on a model that uses multi-headed attention than a similarly-sized model that uses multi-query attention \cite{shazeer2019fast} or grouped-query attention \cite{ainslie2023gqa} in order to reduce the size of the KV cache. However, reducing the KV cache size allows for a larger batch size to fit within GPU memory constraints, which can further increase the speedup of using Hydragen.

As discussed in Section~\ref{sec:batched-inference}, the cost of attention becomes disproportionately high as the batch size grows, since the arithmetic intensity of most transformer operations improve while attention remains memory-bound. Hydragen greatly improves the hardware utilization of attention, making the comparison of attention FLOPs to other model FLOPs more useful when determining the maximum achievable speedup. In several experiments in Section~\ref{sec:exps}, we include a ``No Attention'' baseline that only runs the non-attention components of the transformer in order to establish an upper bound for attainable throughput. 

Another important consideration when predicting the benefits of Hydragen is the relative number of prefix (shared) tokens compared to suffix (unshared) tokens. Since Hydragen makes no optimizations to attention over suffixes, long suffixes can decrease generation throughput. We explore the impact of suffix length on attention speed in Section~\ref{sec:micro}.

\subsection{Implementation}

We implement Hydragen for the Llama family of models \cite{touvron2023llama, touvron2023llama2, rozière2023code}. We highlight that our implementation is simple: we use no custom CUDA code and write Hydragen entirely in PyTorch\footnote{For non-hierarchical inputs, we've also written a Triton kernel for combining softmax denominators.} plus calls to a fast attention primitive. This contrasts with more sophisticated algorithms like PagedAttention, which require bespoke GPU kernels to read from and update the paged KV cache.
We believe that Hydragen's simplicity will allow it to be easily ported to other hardware platforms such as TPUs, which also have hardware dedicated to fast matrix multiplications. 
In our implementation, we use version 2.3.6 of the \verb|flash-attn| package when attending over the prefix, and a Triton kernel from \verb|xformers| when attending over the suffix. The second kernel allows us to have changing sequence lengths in the suffix KV cache across decoding steps while still adhering to the constraints required to use CUDA graphs.

\section{Experiments}
\label{sec:exps}

\subsection{End-To-End Throughput}
\label{sec:e2e}

We benchmark end-to-end LLM throughput in the setting where many completions are sampled from a single prompt. This is a common technique for improving a model's ability at solving math and coding problems \cite{rozière2023code,Li_2022alphacode}. Our benchmarks evaluate Hydragen against four baselines:

\begin{enumerate}
    \item \textbf{FlashAttention:} We perform inference without any shared prefix optimizations, as if all sequences in the batch were fully distinct. We compute full-sequence attention using the Triton kernel that Hydragen uses for suffix attention, and otherwise use the same codebase as Hydragen. This baseline redundantly stores the prefix's keys and values for every sequence in the batch, causing this method to run out of memory quickly.
    \item \textbf{vLLM:} We use version 0.2.7 of the \verb|vllm| package, which uses the PagedAttention algorithm. vLLM avoids redundant storage of the prefix, allowing much larger batch sizes to be tested. Additionally, because of this non-redundant storage, PagedAttention can achieve a higher GPU cache hit rate when reading the prefix, reducing the cost of redundant reads.
    \item \textbf{vLLM without Detokenization:} We disable incremental detokenization in vLLM (accomplished by commenting out one line in the vLLM codebase), which we observed to improve throughput.
    \item \textbf{No Attention:} We skip all self-attention computations in the transformer. This (functionally incorrect) baseline provides a throughput ceiling and helps to illustrate the cost of different attention implementations relative to the rest of the transformer. Note that the query, key, value, and output projections in the attention block are still performed.
\end{enumerate}

\begin{figure}[h]
\centering
\subfigure[]{\includegraphics[width=0.49\textwidth]{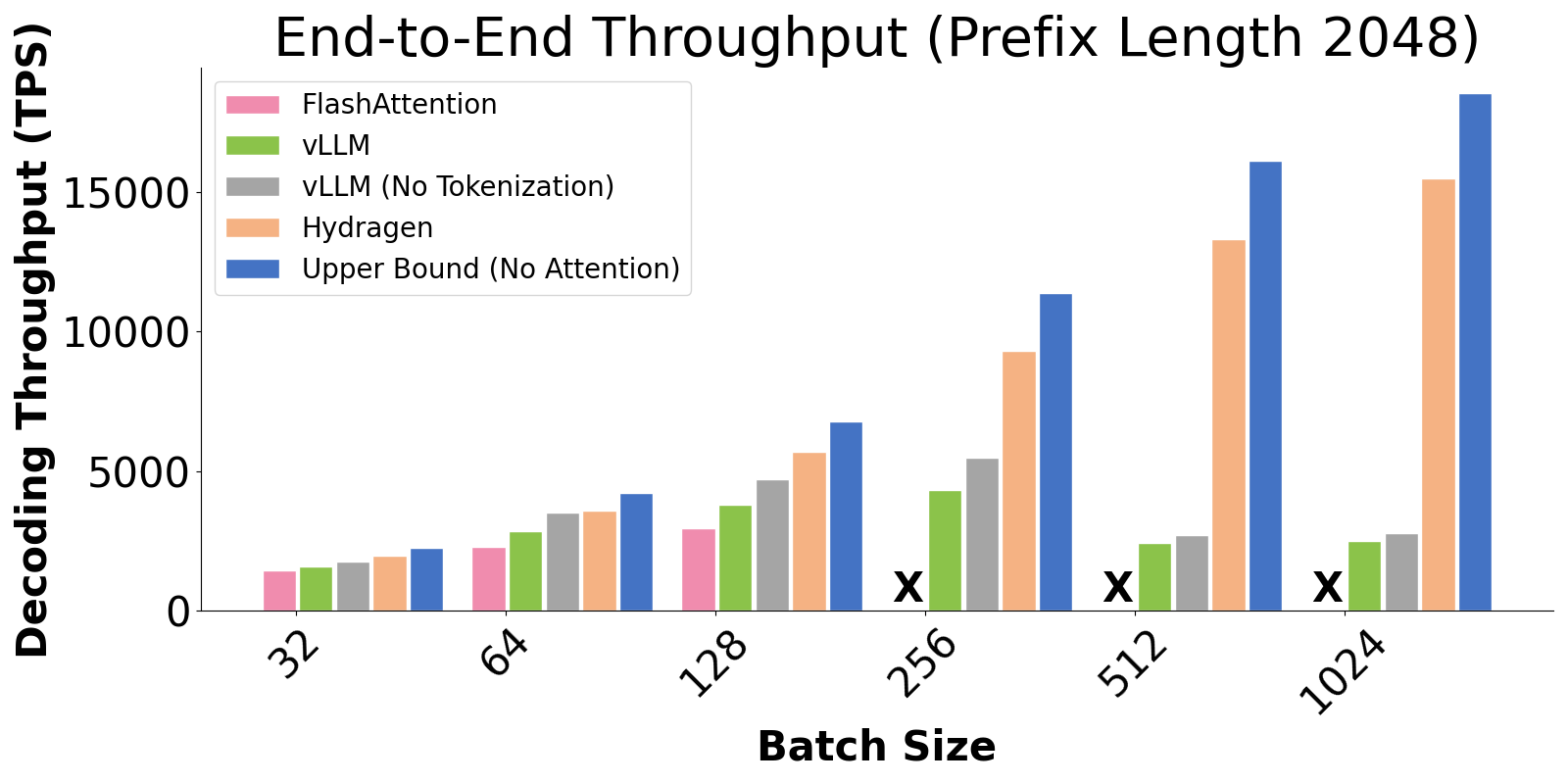}\label{fig:synth-xbs}}
\subfigure[]{\includegraphics[width=0.49\textwidth]{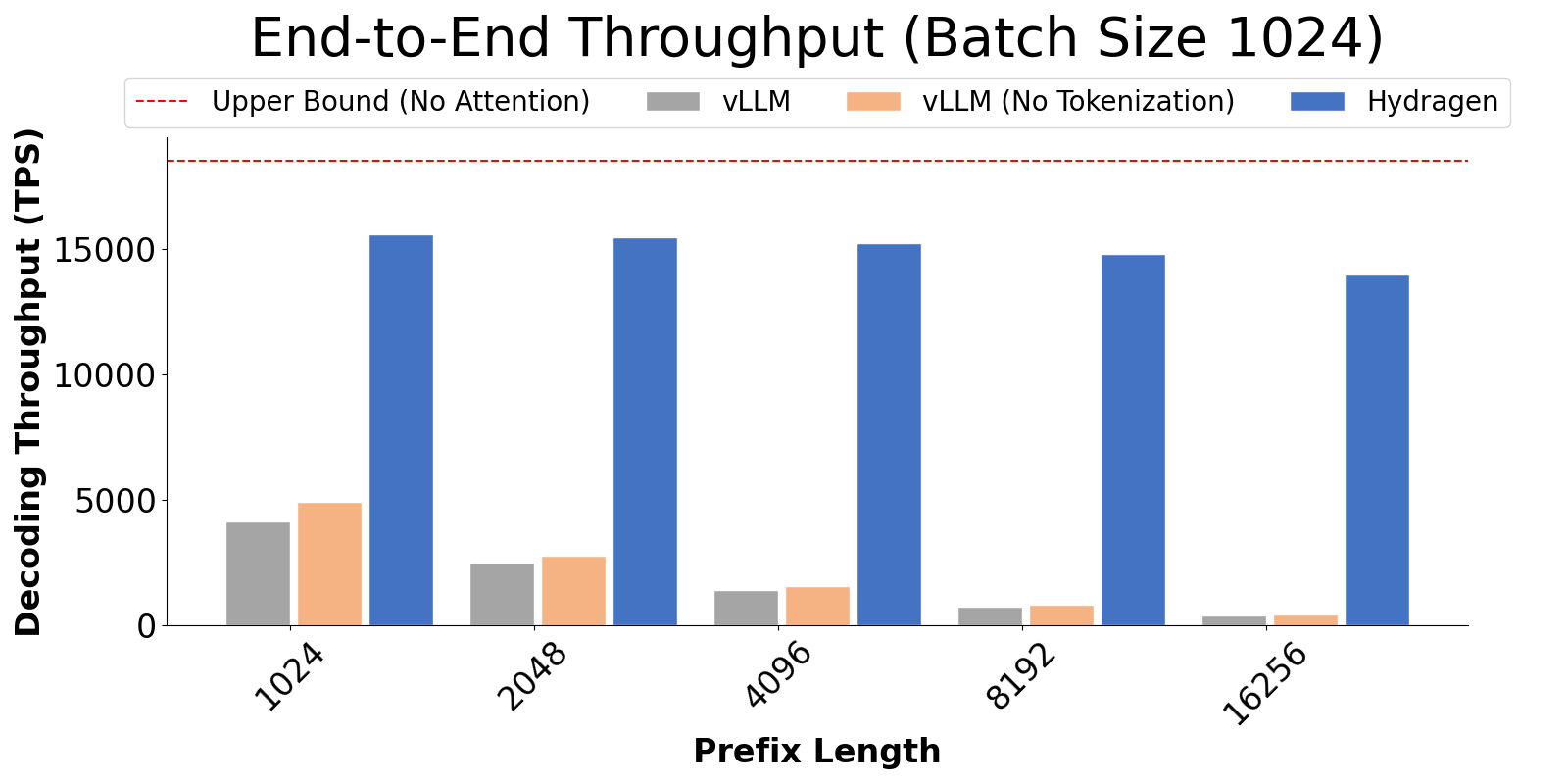}\label{fig:synth-xp}}
\captionof{figure}{Left: End-to-end decoding throughput in tokens per second (TPS) with CodeLlama-13b when generating multiple completions from a prompt containing 2048 tokens. An ``x'' indicates that FlashAttention does not have enough memory to run. As the batch size grows, Hydragen achieves a significantly higher throughput than vLLM baselines. Throughput with Hydragen always remains within 50\% of the upper bound where attention is entirely removed from the model. Details are in Section~\ref{sec:e2e}. Right: Comparing decoding throughput of CodeLlama-13b between Hydragen, vLLM (with and without tokenization), and ``No Attention'', where the attention operation is removed from the model to demonstrate the throughput ceiling. In this scenario where the batch size is fixed, Hydragen improves throughput by up to 32x over the best baseline, with speedups increasing with prefix length.}
\end{figure}

We run our benchmarks on CodeLlama-13b \cite{rozière2023code} and distribute the model with tensor parallelism across eight A100-40GB GPUs in order to have enough GPU memory to store the KV cache with large batch sizes. In Figure~\ref{fig:synth-xbs}, we fix the prefix length to 2048 and sweep over the batch size while generating 128 tokens per completion. When the batch size is small, non-attention operations contribute significantly to decoding time, with all methods reaching at least half of the throughput of no-attention upper bound. At these low batch sizes, Hydragen, the vLLM baselines, and the FlashAttention baselines have similar throughputs. However, as the batch size grows and attention over the prefix becomes increasingly expensive, Hydragen begins to significantly outperform the other baselines. 

In Figure~\ref{fig:synth-xp}, we run a similar experiment, except now we hold the batch size constant at 1024 and sweep over the shared prefix length. The throughput of vLLM significantly decreases as the prefix grows, from just under 5k tokens/second with a prefix length of 1024 to less than 500 tokens/second with a prefix length of 16256. However, with Hydragen, throughput is much less unaffected despite the prefix growing by over 15k tokens. Moreover, across all sequence lengths tested, Hydragen throughput is always within 70\% of the no-attention ceiling. We perform more in-depth sweeps over different models, prefix lengths, batch sizes, and numbers of generated tokens in Appendix~\ref{app:e2e} - for smaller models and shorter completions lengths, Hydragen's speedup can exceed 50x. Additional evaluation setup details are in Appendix~\ref{app:e2e-details}.

\subsection{Microbenchmarking Attention}
\label{sec:micro}

\begin{figure}[h]
    \centering
    \includegraphics[width=0.6\columnwidth]{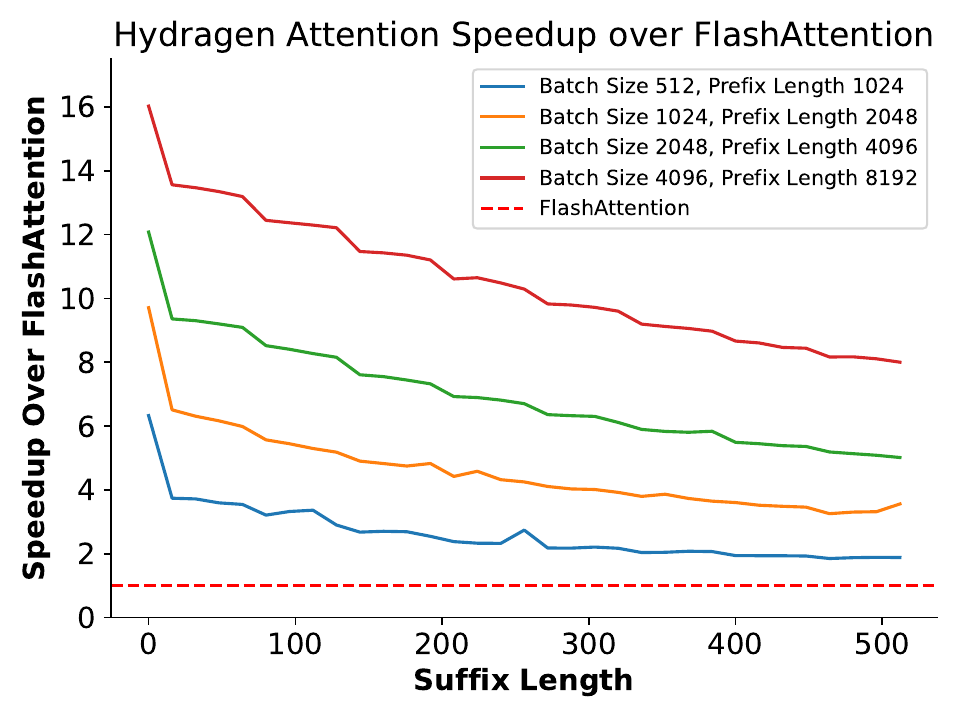}
    \caption{Measuring the speedup of Hydragen attention over FlashAttention across various batch sizes, shared prefix lengths and suffix lengths on a single A100-40GB GPU. We see that Hydragen results in faster inference in all cases, in particular when the ratio of shared length to unique length is high and the batch size is large. We observe even larger performance gains when running on an L40S (a GPU with a higher compute-to-bandwidth ratio than an A100), shown in in Figure~\ref{fig:micro_moregpu}.}

    \label{fig:micro}
\end{figure}

We also perform more granular benchmarks comparing Hydragen attention against FlashAttention, in order to more precisely demonstrate the performance characteristics of our method. Our microbenchmarks run on a single A100-40GB using eight query attention heads, one key and value head, and a head dimension of 128 (matching the setting of CodeLlama-34b when distributed across eight GPUs). We sweep over different batch sizes, prefix lengths, and suffix lengths, reporting our results in Figure~\ref{fig:micro}. Our microbenchmarks corroborate our end-to-end measurements from Section~\ref{sec:e2e} that the speedup with Hydragen increases as the batch size and prefix lengths grow. However, the microbenchmarks also highlight the significant impact of the suffix length on inference time. Hydragen computes attention over suffixes using memory-bound FlashAttention (without inter-sequence batching). As the suffix lengths grow, reading this portion of the KV cache becomes an increasingly significant contributor to total execution time. When generating text using Hydragen, this means that the first tokens decoded by the model are generated the fastest, with throughput decreasing over time as the lengths of completions (and therefore the lengths of suffixes) grow.

These microbenchmarks are also influenced by the hardware platform that they are run on. GPUs with a higher ratio of compute to memory bandwidth benefit more from Hydragen eliminating memory bottlenecks when attending over the prefix. We report results on other GPUs in Appendix~\ref{app:micro} and provide more evaluation details in Appendix~\ref{app:micro-details}.

\subsection{Long Document Question Answering}
\label{sec:long-doc}

\begin{figure}[h]
    \centering
    \includegraphics[width=0.6\columnwidth]{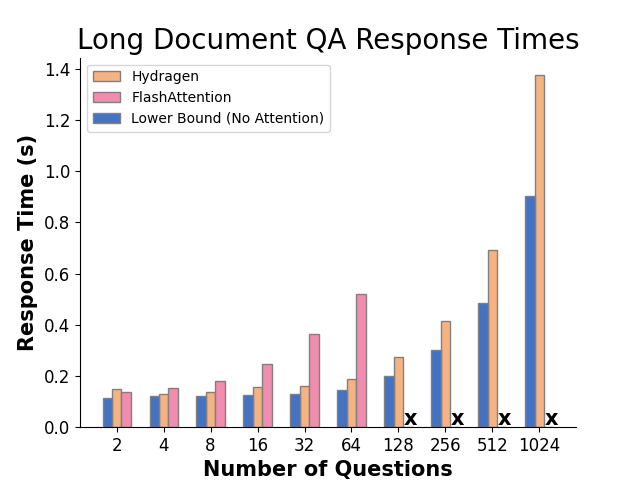}
    \caption{Time to answer questions about a 19947 token-long document when benchmarking Yi-6B-200k on four A100-40GB GPUs. 
    An ``x'' indicates that FlashAttention does not have enough memory to run. Time to process prefix is excluded.}
    \label{fig:needles}
\end{figure}

Additionally, we explore the performance of Hydragen on workloads involving very long documents. We construct a document by embedding synthetic facts into an excerpt of \textit{War and Peace} \cite{Tolstoy1869}. Our shared prefix, totalling 19947 tokens, contains both the document as well as five few-shot examples of question/answer pairs. Our benchmark evaluates Yi-6B-200k \cite{01aiYiModel} on its ability to answer questions based on the embedded facts. We run this benchmark across four A100-40GB GPUs using Hydragen in addition to our FlashAttention and no-attention baselines. Results are reported in Figure~\ref{fig:needles}. We observe that processing time for the FlashAttention baseline rapidly grows far beyond the time of the no-attention baseline, highlighting how attention is the dominant operation for this configuration. Meanwhile, Hydragen's processing time remains within 60\% of the no-attention optimum. Notably, Hydragen can process 256 questions in less time than it takes the FlashAttention baseline to process 64 questions. We provide additional evaluation details in Appendix~\ref{app:needles-details}.

\subsection{Hierarchical Sharing in Competitive Programming}
\label{sec:exps-hierarchy}

\begin{figure}[h]
    \centering
    \includegraphics[width=0.6\columnwidth]{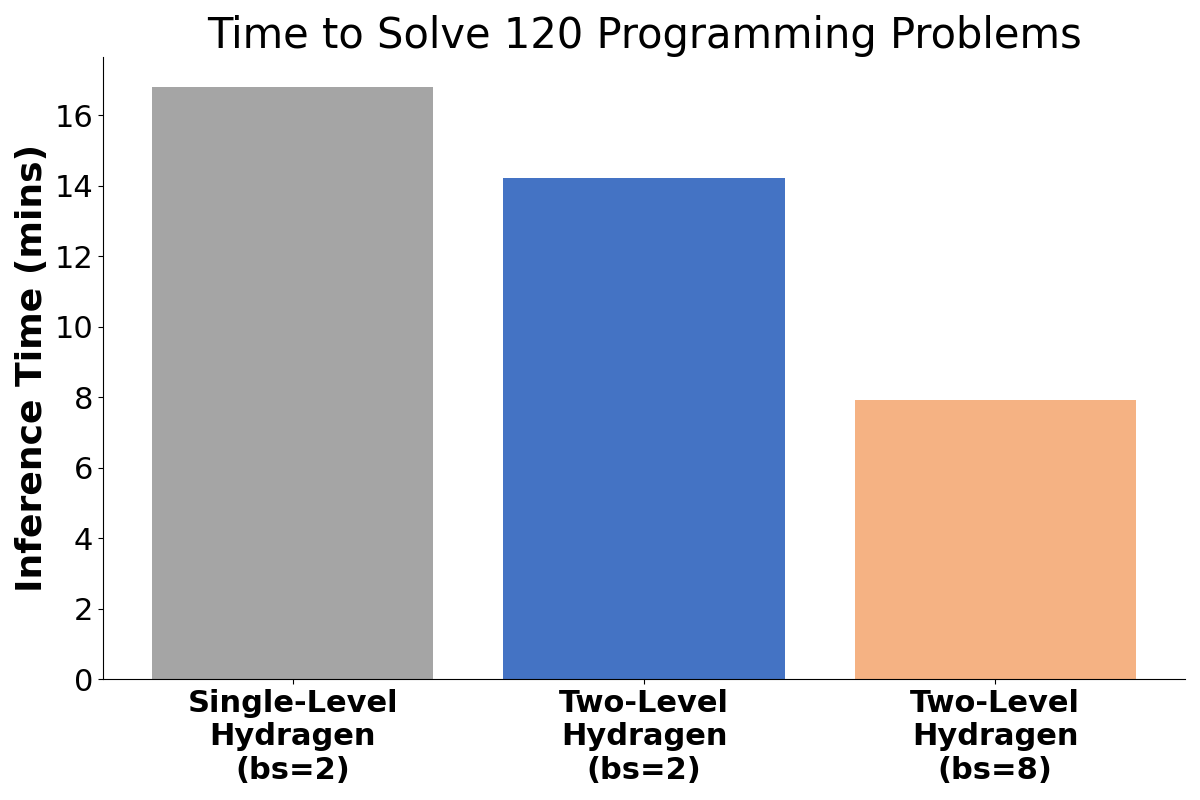}
    \caption{Time to run inference over a dataset of 120 APPS coding problems, sampling 128 solutions per problem with two few-shot examples. The batch size refers to the number of problems processed simultaneously. Across all Hydragen runs (both single and two-level), the few-shot prompt is shared across all sequences. By additionally sharing the problem description across generated candidate solutions, two-level Hydragen decreases overall inference time by an extra 55\% over single-level Hydragen.}
    \label{fig:hierarchy}
\end{figure}

We lastly demonstrate the benefits of applying Hydragen to a setting with hierarchical sharing. Competitive programming was a motivating application for developing our method, since current state-of-the-art systems can sample thousands or more candidate programs from prompts that can contain thousands of tokens~\cite{Li_2022alphacode, rozière2023code}. In this experiment, we benchmark the total time required to evaluate CodeLlama-7b on 120 problems from the APPS dataset~\cite{hendrycksapps2021} using a two-shot prompt and 128 candidate programs per problem. When multiple problems are processed in a single batch, prompt overlap occurs across two levels: the few-shot prompt is shared across all sequences in the batch, while each problem's description is shared across all of the problem's candidate solutions (see Figure~\ref{fig:hierarchy}). 

We run this benchmark using two methods:

\begin{enumerate}
    \item \textbf{Single-Level Hydragen:} We use a single-level version of Hydragen to share the few-shot prompt across all sequences in the batch, and not share problem descriptions across candidate solutions. This leads to redundant storage of the problem description across all candidate solutions, reducing the maximum batch size that can be used.
    \item \textbf{Two-Level Hydragen:} We apply Hydragen across both levels of prompt overlap. This has the dual benefits of improving attention efficiency (by increasing the degree of sharing) as well as avoiding redundant storage, which allows us to increase the batch size used for evaluation. We avoid conflating these benefits by evaluating two-level Hydragen twice: once with the same batch size used for single-level Hydragen, and once with an enlarged batch size.
\end{enumerate}

We report our results in Figure~\ref{fig:hierarchy}. We see that even when the batch size is held constant, adding a second level of sharing to Hydragen can improve attention efficiency and decrease dataset evaluation time by 18\%. Furthermore, the memory saved due to not redundantly storing the problem description allows us to increase the batch size, which in turn results in an additional 45\% reduction in evaluation time. We provide additional evaluation details in Appendix~\ref{app:hierarchy-details}.

\section{Discussion}

In this work we introduced Hydragen, an exact, hardware-aware implementation of attention for batches of sequences that share common prefixes. Our method separates attention over shared prefixes from attention over unique suffixes. This allows us to batch attention queries across sequences when attending over the prefix, reducing redundant memory reads and enabling the use of tensor cores. Hydragen can improve LLM throughput in scenarios where attention is a significant contributor to decoding time, with the greatest speedup occurring when the batch size is large, the shared prefix lengths are long, and the unique suffix lengths are short.

We emphasize that Hydragen is an optimization that can be applied as part of a larger inference framework, and is not intended to be an end-to-end inference solution. Our proof-of-concept implementation of Hydragen requires that the user specifies where sharing occurs across the input sequences. We are excited about future work that incorporates Hydragen into systems that continuously receive requests and schedule sequences for generation \cite{yu2022orca, kwon2023vllm}, such that overlapping sequences can be dynamically identified and exploited.

We hope that our work inspires new LLM algorithms that leverage efficient handling of shared prefixes. Hydragen's ability to significantly expand the shared prefix without a significant throughput penalty should allow models to be provided with much more context than was previously practical. Moreover, we hope that Hydragen's ability to generalize to tree-shaped sharing patterns can assist with research that uses LLMs to explore many possible solutions before deciding on a final output.

\section{Related Work}

\textbf{Transformers and Language Models:} The transformer architecture has enabled significant improvements in state-of-the-art language models \cite{vaswani2023attention}. A defining feature of transformers is that their performance consistently improves when scaling up data and model size \cite{radford2019language, brown2020gpt3, chowdhery2022palm, hoffmann2022chinchilla, openai2023gpt4}. LLM-powered assistants such as ChatGPT have been widely adopted and are currently used by over a hundred million users \cite{OpenAIChatGPT2023}, motivating research into how these models can be deployed more efficiently.

\textbf{KV Cache Management:} Managing large KV caches is a challenge when deploying LLMs. MQA \cite{shazeer2019fast} and GQA \citep{ainslie2023gqa} modify the transformer architecture in order to reduce the KV cache size. These techniques decrease the number of key-value attention heads and assign multiple query heads to a single key-value head. Alternative approaches operate at a systems level, dynamically moving keys and values between GPU memory, CPU memory, and disk \cite{sheng2023flexgen, aminabadi2022deepspeed, huggingfaceAccelerate}. vLLM \cite{kwon2023vllm} introduces a virtual paging system that enables fine-grained KV cache management. This virtual paging can also avoid redundant storage of a prefix's keys and values. Concurrent with our work, SGLang \cite{zheng2023sglang} also investigates and optimizes inference with sequences that have complicated prompt sharing patterns. Their RadixAttention algorithm dynamically scans incoming requests to find the largest subsequence that has already been processed, avoiding the recomputation of overlapping keys and values. Importantly, while both vLLM and RadixAttention avoid redundant storage of overlapping keys and values, they do not optimize the attention computation itself.

\textbf{Hardware-Aware Algorithms:} Algorithms that leverage an understanding of the underlying hardware platform can significantly improve device utilization. Hardware-awareness has significantly improved the efficiency of the attention operation \cite{rabe2022selfattention, dao2022flashattention, dao2023flashattention2}, reducing the memory requirements from $O(N^2)$ to $O(N)$ while improving execution time by avoiding redundant memory transfers. In addition to improving input-output (IO) transfers, many GPU-aware algorithms (including Hydragen) focus on leveraging tensor cores \cite{fu2023flashfftconv}, which can achieve over 10x more FLOPS than the rest of the GPU.

\textbf{LLM Algorithms:} Recent work has demonstrated that LLM capabilities can be improved when many potential solutions are explored when solving a problem. Self-consistency \cite{wang2023selfconsistency} improves performance on arithmetic reasoning tasks by sampling many solutions to a single problem and using a majority-voting protocol. On competitive programming problems, LLMs perform substantially better when many different attempts to a problem are sampled \cite{rozière2023code}. AlphaCode \cite{Li_2022alphacode}, a state-of-the-art competitive programming system, samples as many as a million programs to solve a single problem. Tree-of-Thoughts \cite{yao2023tree} introduces an explicit tree-based search algorithm for solving problems that can be decomposed into discrete decision points. All of these scenarios involve performing batched text generation with overlapping prefixes, which Hydragen is specifically optimized for.

\section{Acknowledgements}

We thank Together AI for partially sponsoring the compute for this project. We also thank Aryaman Arora, Chris Fifty, Jerry Liu, Jon Saad-Falcon, Mayee Chen, Neel Guha, Sabri Eyuboglu, and Vishnu Sarukkai for providing feedback on drafts of this paper.  \\

We gratefully acknowledge the support of NIH under No. U54EB020405 (Mobilize), NSF under Nos. CCF2247015 (Hardware-Aware), CCF1763315 (Beyond Sparsity), CCF1563078 (Volume to Velocity), and 1937301 (RTML); US DEVCOM ARL under Nos. W911NF-23-2-0184 (Long-context) and W911NF-21-2-0251 (Interactive Human-AI Teaming); ONR under Nos. N000142312633 (Deep Signal Processing), N000141712266 (Unifying Weak Supervision), N000142012480 (Non-Euclidean Geometry), and N000142012275 (NEPTUNE); Stanford HAI under No. 247183; NXP, Xilinx, LETI-CEA, Intel, IBM, Microsoft, NEC, Toshiba, TSMC, ARM, Hitachi, BASF, Accenture, Ericsson, Qualcomm, Analog Devices, Google Cloud, Salesforce, Total, the HAI-GCP Cloud Credits for Research program,  the Stanford Data Science Initiative (SDSI), and members of the Stanford DAWN project: Facebook, Google, and VMWare. The U.S. Government is authorized to reproduce and distribute reprints for Governmental purposes notwithstanding any copyright notation thereon. Any opinions, findings, and conclusions or recommendations expressed in this material are those of the authors and do not necessarily reflect the views, policies, or endorsements, either expressed or implied, of NIH, ONR, or the U.S. Government.

\bibliography{main}
\bibliographystyle{plain}

\newpage
\appendix

\section{Proving the Correctness of Attention Decomposition}
\label{app:decomposition}

We start by explicitly expressing softmax as an exponentiation followed by a normalization:

\begin{align}
    \softmax{\frac{QK^T}{\sqrt{d}}} &= \frac{\exp\jjpar{\frac{QK^T}{\sqrt{d}}}}{e^{\LSE{Q, K}}}
\end{align}

Therefore we can rewrite Equation~\ref{eq:sdp} as:

\begin{align}
    \text{SDP}(Q, K, V) &= \jjpar{\frac{\exp\jjpar{\frac{QK^T}{\sqrt{d}}}}{e^{\LSE{Q, K}}}}V
\end{align}

We can then expand Equation~\ref{eq:decomposition}:

\begin{align}
    & \frac{\SDP{Q, K_1, V_1}
 e^{\LSE{Q, K_1}} + \SDP{Q, K_2, V_2}
 e^{\LSE{Q, K_2}}  }{e^{\LSE{Q, K_1}} + e^{\LSE{Q, K_2}}}  \\
 &= \frac{   \jjpar{\frac{\exp\jjpar{\frac{QK_1^T}{\sqrt{d}}}}{e^{\LSE{Q, K_1}}}}V_1
 e^{\LSE{Q, K_1}} + \jjpar{\frac{\exp\jjpar{\frac{QK_2^T}{\sqrt{d}}}}{e^{\LSE{Q, K_2}}}}V_2
 e^{\LSE{Q, K_2}}  }{e^{\LSE{Q, K_1}} + e^{\LSE{Q, K_2}}} \\
 &= \frac{   \exp\jjpar{\frac{QK_1^T}{\sqrt{d}}}V_1
  + \exp\jjpar{\frac{QK_2^T}{\sqrt{d}}}V_2 }{e^{\LSE{Q, K_1}} + e^{\LSE{Q, K_2}}} \\
  &= \frac{   \exp\jjpar{\frac{Q(K_1 || K_2)^T}{\sqrt{d}}}(V_1 || V_2)
   }{e^{\LSE{Q, K_1 || K_2}}} \\
   &= \SDP{Q, K_1 || K_2, V_1 || V_2} 
\end{align}

as required. \qed

\section{Hydragen Pseudocode}
\label{app:code}

We provide PyTorch-style pseudocode implementing Hydragen attention below. We highlight that Hydragen can be implemented easily and efficiently in existing machine learning libraries, as long as there is a fast attention primitive that returns the LSE needed for softmax recombination.

\begin{lstlisting}[language=Python]
import torch
from torch import Tensor

def attention(q: Tensor, k: Tensor, v: Tensor) -> tuple[Tensor, Tensor]:
    """
    Placeholder for some fast attention primitive 
    that also returns LSEs. We use the flash-attn 
    package in our implementation.

    q shape: [batch, qseq_len, qheads, dim]
    k shape: [batch, kvseq_len, kvheads, dim]
    v shape: [batch, kvseq_len, kvheads, dim]
    """
    pass

def combine_lse(
    out1: Tensor,
    lse1: Tensor,
    out2: Tensor,
    lse2: Tensor,
):
    """
    Combines two attention results using their LSEs.

    Out1/2 shape: [batch, seq_len, qheads, hdim]
    lse1/2 shape: [batch, seq_len, qheads]
    """
    max_lse = torch.maximum(lse1, lse2)

    adjust_factor1 = (lse1 - max_lse).exp()
    adjust_factor2 = (lse2 - max_lse).exp()

    new_denominator = adjust_factor1 + adjust_factor2

    aggregated = (
        out1 * adjust_factor1.unsqueeze(-1) + out2 * adjust_factor2.unsqueeze(-1)
    ) / new_denominator.unsqueeze(-1)

    return aggregated


def hydragen_attention(
    q: Tensor,
    prefix_k: Tensor,
    prefix_v: Tensor,
    suffix_k: Tensor,
    suffix_v: Tensor,
):
    """
    q: shape [batch, num_queries (1 during decoding), qheads, dim]

    prefix_k: shape [prefix_len, kvheads, dim]
    prefix_v: shape [prefix_len, kvheads, dim]

    suffix_k: shape [batch, suffix_len, kvheads, dim]
    suffix_v: shape [batch, suffix_len, kvheads, dim]
    """

    b, nq, hq, d = q.shape

    # inter-sequence batching: merge attention queries
    # as if they all came from the same sequence.
    batched_q = q.view(1, b * nq, hq, d)


    # efficient attention over prefixes
    # prefix_out: shape [1, batch * nq, hq, dim]
    # prefix_lse: shape [1, batch * nq, hq]
    prefix_out, prefix_lse = attention(
        batched_q,
        prefix_k.unsqueeze(0),
        prefix_v.unsqueeze(0),
    )


    # normal attention over suffixes
    # suffix_out: shape [batch, suffix_len, hq, dim]
    # suffix_lse: shape [batch, suffix_len, hq]
    suffix_out, suffix_lse = attention(
        batched_q,
        suffix_k,
        suffix_v,
    )

    # unmerge prefix attention results and combine
    # softmax denominators
    aggregated = combine_lse(
        prefix_out.view(b, nq, hq, d),
        prefix_lse.view(b, nq, hq),
        suffix_out,
        suffix_lse,
    )

    return aggregated
\end{lstlisting}

\section{Additional Results}

\subsection{End-to-End Throughput}
\label{app:e2e}

We expand on the end-to-end throughput experiments discussed in Section~\ref{sec:e2e}. We report additional results with more model sizes when generating 128 and 256 tokens. These results are displayed in Table~\ref{tab:fsynth-7b} and Table~\ref{tab:fsynth-7b-c256} for CodeLlama-7b, Table~\ref{tab:fsynth-13b} and Table~\ref{tab:fsynth-13b-c256} for CodeLlama-13b, and Table~\ref{tab:fsynth-34b} and Table~\ref{tab:fsynth-34b-c256} for CodeLlama-34b, respectively \cite{rozière2023code}. Note that in the tables where 128 tokens are generated per sequence, the ``16K'' column corresponds to a prefix length of 16256 tokens, while for the tables with 256 generated tokens per sequence, this corresponds to 16128 tokens (this is done to accommodate the 16384 max sequence length of the CodeLlama models).

    \begin{table}[H]\centering\tiny\noindent\makebox[\textwidth]{\begin{tabular}{|m{0.33cm}||m{.25cm}|m{.25cm}|m{.25cm}|m{.25cm}|m{.25cm}||m{.25cm}|m{.25cm}|m{.25cm}|m{.25cm}|m{.25cm}||m{.25cm}|m{.25cm}|m{.25cm}|m{.25cm}|m{.25cm}||m{.25cm}|m{.25cm}|m{.25cm}|m{.25cm}|m{.25cm}||>{\centering\arraybackslash}m{3cm}|}
    \hline & \multicolumn{5}{c||}{FlashAttention} & \multicolumn{5}{c||}{Hydragen} & \multicolumn{5}{c||}{vLLM (No Tokenization)} & \multicolumn{5}{c||}{vLLM} & \multicolumn{1}{c|}{Upper Bound (No Attention)}  \\
    \hline Batch & \multicolumn{5}{c||}{Prefix length} & \multicolumn{5}{c||}{Prefix length} & \multicolumn{5}{c||}{Prefix length} & \multicolumn{5}{c||}{Prefix length} & \multicolumn{1}{c|}{Prefix length}  \\
    \cline{2-22} Size & \vspace{0.02cm} $1$K\vspace{0.02cm} & \vspace{0.02cm} $2$K\vspace{0.02cm} & \vspace{0.02cm} $4$K\vspace{0.02cm} & \vspace{0.02cm} $8$K\vspace{0.02cm} & \vspace{0.02cm} $16$K\vspace{0.02cm} & \vspace{0.02cm} $1$K\vspace{0.02cm} & \vspace{0.02cm} $2$K\vspace{0.02cm} & \vspace{0.02cm} $4$K\vspace{0.02cm} & \vspace{0.02cm} $8$K\vspace{0.02cm} & \vspace{0.02cm} $16$K\vspace{0.02cm} & \vspace{0.02cm} $1$K\vspace{0.02cm} & \vspace{0.02cm} $2$K\vspace{0.02cm} & \vspace{0.02cm} $4$K\vspace{0.02cm} & \vspace{0.02cm} $8$K\vspace{0.02cm} & \vspace{0.02cm} $16$K\vspace{0.02cm} & \vspace{0.02cm} $1$K\vspace{0.02cm} & \vspace{0.02cm} $2$K\vspace{0.02cm} & \vspace{0.02cm} $4$K\vspace{0.02cm} & \vspace{0.02cm} $8$K\vspace{0.02cm} & \vspace{0.02cm} $16$K\vspace{0.02cm}& All \\
    \hline 32 & 2.5\newline$\pm$\newline0.0 & 2.2\newline$\pm$\newline0.0 & 1.8\newline$\pm$\newline0.0 & 1.3\newline$\pm$\newline0.0 & 0.9\newline$\pm$\newline0.0 & 2.7\newline$\pm$\newline0.0 & 2.7\newline$\pm$\newline0.0 & 2.6\newline$\pm$\newline0.0 & 2.6\newline$\pm$\newline0.0 & 2.5\newline$\pm$\newline0.0 & 1.7\newline$\pm$\newline0.0 & 1.8\newline$\pm$\newline0.0 & 1.7\newline$\pm$\newline0.1 & 0.6\newline$\pm$\newline0.0 & 0.4\newline$\pm$\newline0.0 & 1.6\newline$\pm$\newline0.0 & 1.6\newline$\pm$\newline0.0 & 1.5\newline$\pm$\newline0.0 & 0.6\newline$\pm$\newline0.0 & 0.3\newline$\pm$\newline0.0 & $3.1\pm0.0$\\
 \hline64 & 4.2\newline$\pm$\newline0.0 & 3.4\newline$\pm$\newline0.0 & 2.6\newline$\pm$\newline0.0 & 1.7\newline$\pm$\newline0.0 & X & 5.0\newline$\pm$\newline0.0 & 4.9\newline$\pm$\newline0.0 & 4.9\newline$\pm$\newline0.1 & 4.8\newline$\pm$\newline0.0 & 4.6\newline$\pm$\newline0.0 & 3.5\newline$\pm$\newline0.1 & 3.5\newline$\pm$\newline0.1 & 2.9\newline$\pm$\newline0.1 & 0.7\newline$\pm$\newline0.0 & 0.4\newline$\pm$\newline0.0 & 2.9\newline$\pm$\newline0.0 & 2.8\newline$\pm$\newline0.1 & 2.1\newline$\pm$\newline0.2 & 0.7\newline$\pm$\newline0.0 & 0.4\newline$\pm$\newline0.0 & $5.7\pm0.0$\\
 \hline128 & 5.7\newline$\pm$\newline0.0 & 4.2\newline$\pm$\newline0.0 & 2.7\newline$\pm$\newline0.0 & X & X & 8.6\newline$\pm$\newline0.0 & 8.5\newline$\pm$\newline0.0 & 8.4\newline$\pm$\newline0.0 & 8.3\newline$\pm$\newline0.0 & 8.0\newline$\pm$\newline0.0 & 6.1 & 5.5 & 3.2 & 0.8 & 0.4 & 4.9 & 4.5 & 2.7 & 0.7 & 0.4 & $10.3\pm0.0$\\
 \hline256 & 8.1\newline$\pm$\newline0.0 & 5.7\newline$\pm$\newline0.0 & X & X & X & 13.3\newline$\pm$\newline0.0 & 13.3\newline$\pm$\newline0.0 & 13.1\newline$\pm$\newline0.0 & 12.8\newline$\pm$\newline0.0 & 12.3\newline$\pm$\newline0.0 & 8.9 & 5.6 & 3.1 & 0.8 & 0.4 & 6.9 & 4.2 & 2.5 & 0.8 & 0.4 & $15.8\pm0.0$\\
 \hline512 & X & X & X & X & X & 19.6\newline$\pm$\newline0.0 & 19.4\newline$\pm$\newline0.0 & 19.1\newline$\pm$\newline0.0 & 18.5\newline$\pm$\newline0.0 & 17.5\newline$\pm$\newline0.0 & 4.7 & 2.8 & 1.5 & 0.8 & 0.4 & 4.2 & 2.5 & 1.4 & 0.8 & 0.4 & $23.2\pm0.0$\\
 \hline1024 & X & X & X & X & X & 25.3\newline$\pm$\newline0.0 & 25.1\newline$\pm$\newline0.0 & 24.7\newline$\pm$\newline0.0 & 23.9\newline$\pm$\newline0.0 & 22.4\newline$\pm$\newline0.0 & 4.9 & 2.8 & 1.5 & 0.8 & 0.4 & 4.2 & 2.5 & 1.4 & 0.7 & 0.4 & $30.1\pm0.0$\\
 \hline2048 & X & X & X & X & X & 27.9\newline$\pm$\newline0.0 & 27.5\newline$\pm$\newline0.0 & 26.7\newline$\pm$\newline0.0 & 25.3\newline$\pm$\newline0.0 & 22.8\newline$\pm$\newline0.0 & 4.9 & 2.8 & 1.5 & 0.8 & 0.4 & 4.2 & 2.5 & 1.4 & 0.7 & 0.4 & $32.9\pm0.0$\\ \hline
    \end{tabular}}
    \caption{End-to-end decoding throughput (thousands of tokens per second) with CodeLlama-7B on 8xA100 40 GB GPUs when generating 128 tokens. An x indicates the model does not have the required memory to run.}
    \label{tab:fsynth-7b}
    \end{table}

    \begin{table}[H]\centering\tiny\noindent\makebox[\textwidth]{\begin{tabular}{|m{0.33cm}||m{.25cm}|m{.25cm}|m{.25cm}|m{.25cm}|m{.25cm}||m{.25cm}|m{.25cm}|m{.25cm}|m{.25cm}|m{.25cm}||m{.25cm}|m{.25cm}|m{.25cm}|m{.25cm}|m{.25cm}||m{.25cm}|m{.25cm}|m{.25cm}|m{.25cm}|m{.25cm}||>{\centering\arraybackslash}m{3cm}|}
    \hline & \multicolumn{5}{c||}{FlashAttention} & \multicolumn{5}{c||}{Hydragen} & \multicolumn{5}{c||}{vLLM (No Tokenization)} & \multicolumn{5}{c||}{vLLM} & \multicolumn{1}{c|}{Upper Bound (No Attention)}  \\
    \hline Batch & \multicolumn{5}{c||}{Prefix length} & \multicolumn{5}{c||}{Prefix length} & \multicolumn{5}{c||}{Prefix length} & \multicolumn{5}{c||}{Prefix length} & \multicolumn{1}{c|}{Prefix length}  \\
    \cline{2-22} Size & \vspace{0.02cm} $1$K\vspace{0.02cm} & \vspace{0.02cm} $2$K\vspace{0.02cm} & \vspace{0.02cm} $4$K\vspace{0.02cm} & \vspace{0.02cm} $8$K\vspace{0.02cm} & \vspace{0.02cm} $16$K\vspace{0.02cm} & \vspace{0.02cm} $1$K\vspace{0.02cm} & \vspace{0.02cm} $2$K\vspace{0.02cm} & \vspace{0.02cm} $4$K\vspace{0.02cm} & \vspace{0.02cm} $8$K\vspace{0.02cm} & \vspace{0.02cm} $16$K\vspace{0.02cm} & \vspace{0.02cm} $1$K\vspace{0.02cm} & \vspace{0.02cm} $2$K\vspace{0.02cm} & \vspace{0.02cm} $4$K\vspace{0.02cm} & \vspace{0.02cm} $8$K\vspace{0.02cm} & \vspace{0.02cm} $16$K\vspace{0.02cm} & \vspace{0.02cm} $1$K\vspace{0.02cm} & \vspace{0.02cm} $2$K\vspace{0.02cm} & \vspace{0.02cm} $4$K\vspace{0.02cm} & \vspace{0.02cm} $8$K\vspace{0.02cm} & \vspace{0.02cm} $16$K\vspace{0.02cm}& All \\
    \hline 32 & 2.4\newline$\pm$\newline0.0 & 2.2\newline$\pm$\newline0.0 & 1.8\newline$\pm$\newline0.0 & 1.3\newline$\pm$\newline0.0 & 0.9\newline$\pm$\newline0.0 & 2.6\newline$\pm$\newline0.0 & 2.6\newline$\pm$\newline0.0 & 2.6\newline$\pm$\newline0.0 & 2.5\newline$\pm$\newline0.0 & 2.4\newline$\pm$\newline0.0 & 1.7\newline$\pm$\newline0.0 & 1.8\newline$\pm$\newline0.0 & 1.7\newline$\pm$\newline0.0 & 0.6\newline$\pm$\newline0.0 & 0.4\newline$\pm$\newline0.0 & 1.6\newline$\pm$\newline0.0 & 1.5\newline$\pm$\newline0.0 & 1.5\newline$\pm$\newline0.0 & 0.6\newline$\pm$\newline0.0 & 0.3\newline$\pm$\newline0.0 & $3.1\pm0.0$\\
 \hline64 & 3.9\newline$\pm$\newline0.0 & 3.4\newline$\pm$\newline0.0 & 2.5\newline$\pm$\newline0.0 & 1.7\newline$\pm$\newline0.0 & X & 4.8\newline$\pm$\newline0.0 & 4.8\newline$\pm$\newline0.0 & 4.8\newline$\pm$\newline0.0 & 4.7\newline$\pm$\newline0.0 & 4.5\newline$\pm$\newline0.0 & 3.4\newline$\pm$\newline0.0 & 3.3\newline$\pm$\newline0.0 & 2.7\newline$\pm$\newline0.0 & 0.7\newline$\pm$\newline0.0 & 0.4\newline$\pm$\newline0.0 & 2.8\newline$\pm$\newline0.1 & 2.8\newline$\pm$\newline0.0 & 2.3\newline$\pm$\newline0.0 & 0.6\newline$\pm$\newline0.0 & 0.4\newline$\pm$\newline0.0 & $5.7\pm0.0$\\
 \hline128 & 5.3\newline$\pm$\newline0.0 & 4.1\newline$\pm$\newline0.0 & 2.7\newline$\pm$\newline0.0 & X & X & 8.2\newline$\pm$\newline0.0 & 8.2\newline$\pm$\newline0.0 & 8.1\newline$\pm$\newline0.0 & 7.9\newline$\pm$\newline0.0 & 7.7\newline$\pm$\newline0.0 & 6.3 & 5.0 & 2.9 & 0.8 & 0.4 & 4.8 & 4.0 & 2.5 & 0.7 & 0.4 & $10.2\pm0.0$\\
 \hline256 & 7.4\newline$\pm$\newline0.0 & X & X & X & X & 12.7\newline$\pm$\newline0.0 & 12.6\newline$\pm$\newline0.0 & 12.5\newline$\pm$\newline0.0 & 12.2\newline$\pm$\newline0.0 & 11.8\newline$\pm$\newline0.0 & 8.8 & 5.5 & 3.1 & 0.8 & 0.4 & 6.5 & 4.2 & 2.5 & 0.7 & 0.4 & $15.7\pm0.0$\\
 \hline512 & X & X & X & X & X & 18.4\newline$\pm$\newline0.0 & 18.2\newline$\pm$\newline0.0 & 18.0\newline$\pm$\newline0.0 & 17.5\newline$\pm$\newline0.0 & 16.6\newline$\pm$\newline0.0 & 4.6 & 2.8 & 1.6 & 0.8 & 0.4 & 3.8 & 2.4 & 1.4 & 0.7 & 0.4 & $23.2\pm0.0$\\
 \hline1024 & X & X & X & X & X & 23.4\newline$\pm$\newline0.0 & 23.2\newline$\pm$\newline0.0 & 22.9\newline$\pm$\newline0.0 & 22.2\newline$\pm$\newline0.0 & 21.0\newline$\pm$\newline0.0 & 4.8 & 2.8 & 1.6 & 0.8 & 0.4 & 3.9 & 2.4 & 1.4 & 0.7 & 0.4 & $30.0\pm0.0$\\ \hline
    \end{tabular}}
    \caption{End-to-end decoding throughput (thousands of tokens per second) with CodeLlama-7B on 8xA100 40 GB GPUs when generating 256 tokens. An x indicates the model does not have the required memory to run.}
    \label{tab:fsynth-7b-c256}
    \end{table}

    \begin{table}[H]\centering\tiny\noindent\makebox[\textwidth]{\begin{tabular}{|m{0.33cm}||m{.25cm}|m{.25cm}|m{.25cm}|m{.25cm}|m{.25cm}||m{.25cm}|m{.25cm}|m{.25cm}|m{.25cm}|m{.25cm}||m{.25cm}|m{.25cm}|m{.25cm}|m{.25cm}|m{.25cm}||m{.25cm}|m{.25cm}|m{.25cm}|m{.25cm}|m{.25cm}||>{\centering\arraybackslash}m{3cm}|}
    \hline & \multicolumn{5}{c||}{FlashAttention} & \multicolumn{5}{c||}{Hydragen} & \multicolumn{5}{c||}{vLLM (No Tokenization)} & \multicolumn{5}{c||}{vLLM} & \multicolumn{1}{c|}{Upper Bound (No Attention)}  \\
    \hline Batch & \multicolumn{5}{c||}{Prefix length} & \multicolumn{5}{c||}{Prefix length} & \multicolumn{5}{c||}{Prefix length} & \multicolumn{5}{c||}{Prefix length} & \multicolumn{1}{c|}{Prefix length}  \\
    \cline{2-22} Size & \vspace{0.02cm} $1$K\vspace{0.02cm} & \vspace{0.02cm} $2$K\vspace{0.02cm} & \vspace{0.02cm} $4$K\vspace{0.02cm} & \vspace{0.02cm} $8$K\vspace{0.02cm} & \vspace{0.02cm} $16$K\vspace{0.02cm} & \vspace{0.02cm} $1$K\vspace{0.02cm} & \vspace{0.02cm} $2$K\vspace{0.02cm} & \vspace{0.02cm} $4$K\vspace{0.02cm} & \vspace{0.02cm} $8$K\vspace{0.02cm} & \vspace{0.02cm} $16$K\vspace{0.02cm} & \vspace{0.02cm} $1$K\vspace{0.02cm} & \vspace{0.02cm} $2$K\vspace{0.02cm} & \vspace{0.02cm} $4$K\vspace{0.02cm} & \vspace{0.02cm} $8$K\vspace{0.02cm} & \vspace{0.02cm} $16$K\vspace{0.02cm} & \vspace{0.02cm} $1$K\vspace{0.02cm} & \vspace{0.02cm} $2$K\vspace{0.02cm} & \vspace{0.02cm} $4$K\vspace{0.02cm} & \vspace{0.02cm} $8$K\vspace{0.02cm} & \vspace{0.02cm} $16$K\vspace{0.02cm}& All \\
    \hline 32 & 1.7\newline$\pm$\newline0.0 & 1.4\newline$\pm$\newline0.0 & 1.1\newline$\pm$\newline0.0 & 0.7\newline$\pm$\newline0.0 & X & 2.0\newline$\pm$\newline0.0 & 2.0\newline$\pm$\newline0.0 & 1.9\newline$\pm$\newline0.0 & 1.8\newline$\pm$\newline0.0 & 1.8\newline$\pm$\newline0.0 & 1.8\newline$\pm$\newline0.0 & 1.8\newline$\pm$\newline0.0 & 1.8\newline$\pm$\newline0.0 & 0.6\newline$\pm$\newline0.0 & 0.4\newline$\pm$\newline0.0 & 1.6\newline$\pm$\newline0.0 & 1.6\newline$\pm$\newline0.0 & 1.5\newline$\pm$\newline0.0 & 0.5\newline$\pm$\newline0.0 & 0.3\newline$\pm$\newline0.0 & $2.3\pm0.0$\\
 \hline64 & 2.9\newline$\pm$\newline0.0 & 2.3\newline$\pm$\newline0.0 & 1.6\newline$\pm$\newline0.0 & X & X & 3.6\newline$\pm$\newline0.0 & 3.6\newline$\pm$\newline0.0 & 3.6\newline$\pm$\newline0.0 & 3.4\newline$\pm$\newline0.0 & 3.4\newline$\pm$\newline0.0 & 3.5\newline$\pm$\newline0.1 & 3.5\newline$\pm$\newline0.0 & 2.9\newline$\pm$\newline0.1 & 0.7\newline$\pm$\newline0.0 & 0.4\newline$\pm$\newline0.0 & 3.0\newline$\pm$\newline0.0 & 2.9\newline$\pm$\newline0.1 & 2.4\newline$\pm$\newline0.0 & 0.6\newline$\pm$\newline0.0 & 0.4\newline$\pm$\newline0.0 & $4.2\pm0.0$\\
 \hline128 & 4.0\newline$\pm$\newline0.0 & 2.9\newline$\pm$\newline0.0 & X & X & X & 5.8\newline$\pm$\newline0.0 & 5.7\newline$\pm$\newline0.2 & 5.6\newline$\pm$\newline0.0 & 5.6\newline$\pm$\newline0.0 & 5.7\newline$\pm$\newline0.0 & 5.5 & 4.7\newline$\pm$\newline0.1 & 3.0 & 0.8 & 0.4 & 4.8 & 3.8\newline$\pm$\newline0.1 & 2.6 & 0.7 & 0.4 & $6.8\pm0.0$\\
 \hline256 & 5.7\newline$\pm$\newline0.0 & X & X & X & X & 9.6\newline$\pm$\newline0.0 & 9.3\newline$\pm$\newline0.0 & 9.4\newline$\pm$\newline0.0 & 9.2\newline$\pm$\newline0.0 & 8.8\newline$\pm$\newline0.0 & 8.0 & 5.5\newline$\pm$\newline0.1 & 3.2 & 0.8 & 0.4 & 6.1 & 4.3\newline$\pm$\newline0.1 & 2.7 & 0.7 & 0.4 & $11.4\pm0.0$\\
 \hline512 & X & X & X & X & X & 13.4\newline$\pm$\newline0.0 & 13.3\newline$\pm$\newline0.0 & 13.2\newline$\pm$\newline0.0 & 12.9\newline$\pm$\newline0.0 & 12.3\newline$\pm$\newline0.0 & 4.7 & 2.7\newline$\pm$\newline0.0 & 1.6 & 0.8 & 0.4 & 4.1 & 2.4\newline$\pm$\newline0.0 & 1.4 & 0.8 & 0.4 & $16.1\pm0.0$\\
 \hline1024 & X & X & X & X & X & 15.6\newline$\pm$\newline0.0 & 15.5\newline$\pm$\newline0.0 & 15.3\newline$\pm$\newline0.0 & 14.8\newline$\pm$\newline0.0 & 14.0\newline$\pm$\newline0.0 & 4.9\newline$\pm$\newline0.0 & 2.8\newline$\pm$\newline0.0 & 1.6\newline$\pm$\newline0.0 & 0.8\newline$\pm$\newline0.0 & 0.4\newline$\pm$\newline0.0 & 4.2\newline$\pm$\newline0.0 & 2.5\newline$\pm$\newline0.0 & 1.4\newline$\pm$\newline0.0 & 0.7\newline$\pm$\newline0.0 & 0.4\newline$\pm$\newline0.0 & $18.5\pm0.0$\\ \hline
    \end{tabular}}
    \caption{End-to-end decoding throughput (thousands of tokens per second) with CodeLlama-13B on 8xA100 40 GB GPUs when generating 128 tokens. An x indicates the model does not have the required memory to run.}
    \label{tab:fsynth-13b}
    \end{table}

    \begin{table}[H]\centering\tiny\noindent\makebox[\textwidth]{\begin{tabular}{|m{0.33cm}||m{.25cm}|m{.25cm}|m{.25cm}|m{.25cm}|m{.25cm}||m{.25cm}|m{.25cm}|m{.25cm}|m{.25cm}|m{.25cm}||m{.25cm}|m{.25cm}|m{.25cm}|m{.25cm}|m{.25cm}||m{.25cm}|m{.25cm}|m{.25cm}|m{.25cm}|m{.25cm}||>{\centering\arraybackslash}m{3cm}|}
    \hline & \multicolumn{5}{c||}{FlashAttention} & \multicolumn{5}{c||}{Hydragen} & \multicolumn{5}{c||}{vLLM (No Tokenization)} & \multicolumn{5}{c||}{vLLM} & \multicolumn{1}{c|}{Upper Bound (No Attention)}  \\
    \hline Batch & \multicolumn{5}{c||}{Prefix length} & \multicolumn{5}{c||}{Prefix length} & \multicolumn{5}{c||}{Prefix length} & \multicolumn{5}{c||}{Prefix length} & \multicolumn{1}{c|}{Prefix length}  \\
    \cline{2-22} Size & \vspace{0.02cm} $1$K\vspace{0.02cm} & \vspace{0.02cm} $2$K\vspace{0.02cm} & \vspace{0.02cm} $4$K\vspace{0.02cm} & \vspace{0.02cm} $8$K\vspace{0.02cm} & \vspace{0.02cm} $16$K\vspace{0.02cm} & \vspace{0.02cm} $1$K\vspace{0.02cm} & \vspace{0.02cm} $2$K\vspace{0.02cm} & \vspace{0.02cm} $4$K\vspace{0.02cm} & \vspace{0.02cm} $8$K\vspace{0.02cm} & \vspace{0.02cm} $16$K\vspace{0.02cm} & \vspace{0.02cm} $1$K\vspace{0.02cm} & \vspace{0.02cm} $2$K\vspace{0.02cm} & \vspace{0.02cm} $4$K\vspace{0.02cm} & \vspace{0.02cm} $8$K\vspace{0.02cm} & \vspace{0.02cm} $16$K\vspace{0.02cm} & \vspace{0.02cm} $1$K\vspace{0.02cm} & \vspace{0.02cm} $2$K\vspace{0.02cm} & \vspace{0.02cm} $4$K\vspace{0.02cm} & \vspace{0.02cm} $8$K\vspace{0.02cm} & \vspace{0.02cm} $16$K\vspace{0.02cm}& All \\
    \hline 32 & 1.7\newline$\pm$\newline0.0 & 1.4\newline$\pm$\newline0.0 & 1.1\newline$\pm$\newline0.0 & 0.7\newline$\pm$\newline0.0 & X & 1.9\newline$\pm$\newline0.0 & 1.9\newline$\pm$\newline0.0 & 1.9\newline$\pm$\newline0.0 & 1.8\newline$\pm$\newline0.0 & 1.8\newline$\pm$\newline0.0 & 1.8\newline$\pm$\newline0.0 & 1.7\newline$\pm$\newline0.0 & 1.8\newline$\pm$\newline0.0 & 0.5\newline$\pm$\newline0.0 & 0.3\newline$\pm$\newline0.0 & 1.6\newline$\pm$\newline0.0 & 1.6\newline$\pm$\newline0.0 & 1.5\newline$\pm$\newline0.0 & 0.5\newline$\pm$\newline0.0 & 0.3\newline$\pm$\newline0.0 & $2.3\pm0.0$\\
 \hline64 & 2.8\newline$\pm$\newline0.0 & 2.2\newline$\pm$\newline0.0 & 1.6\newline$\pm$\newline0.0 & X & X & 3.5\newline$\pm$\newline0.0 & 3.5\newline$\pm$\newline0.0 & 3.4\newline$\pm$\newline0.0 & 3.2\newline$\pm$\newline0.0 & 3.3\newline$\pm$\newline0.0 & 3.4\newline$\pm$\newline0.1 & 3.4\newline$\pm$\newline0.0 & 2.9\newline$\pm$\newline0.0 & 0.7\newline$\pm$\newline0.0 & 0.4\newline$\pm$\newline0.0 & 3.0\newline$\pm$\newline0.1 & 2.7\newline$\pm$\newline0.2 & 2.2\newline$\pm$\newline0.1 & 0.6\newline$\pm$\newline0.0 & 0.4\newline$\pm$\newline0.0 & $4.2\pm0.0$\\
 \hline128 & 3.8\newline$\pm$\newline0.0 & 2.8\newline$\pm$\newline0.0 & X & X & X & 5.6\newline$\pm$\newline0.0 & 5.5\newline$\pm$\newline0.0 & 5.3\newline$\pm$\newline0.0 & 5.4\newline$\pm$\newline0.0 & 5.2\newline$\pm$\newline0.0 & 5.4 & 4.6 & 3.0 & 0.8 & 0.4 & 4.6 & 3.7 & 2.4 & 0.7 & 0.4 & $6.8\pm0.0$\\
 \hline256 & 5.4\newline$\pm$\newline0.0 & X & X & X & X & 8.9\newline$\pm$\newline0.0 & 8.7\newline$\pm$\newline0.0 & 8.8\newline$\pm$\newline0.0 & 8.7\newline$\pm$\newline0.0 & 8.4\newline$\pm$\newline0.0 & 7.6 & 5.5 & 3.1 & 0.8 & 0.4 & 5.9 & 4.3 & 2.5 & 0.7 & 0.4 & $11.3\pm0.0$\\
 \hline512 & X & X & X & X & X & 12.3\newline$\pm$\newline0.0 & 12.3\newline$\pm$\newline0.0 & 12.2\newline$\pm$\newline0.0 & 12.0\newline$\pm$\newline0.0 & 11.4\newline$\pm$\newline0.0 & 4.4 & 2.7 & 1.5 & 0.8 & 0.4 & 3.8 & 2.4 & 1.4 & 0.7 & 0.4 & $16.1\pm0.0$\\ \hline
    \end{tabular}}
    \caption{End-to-end decoding throughput (thousands of tokens per second) with CodeLlama-13B on 8xA100 40 GB GPUs when generating 256 tokens. An x indicates the model does not have the required memory to run.}
    \label{tab:fsynth-13b-c256}
    \end{table}

    \begin{table}[H]\centering\tiny\noindent\makebox[\textwidth]{\begin{tabular}{|m{0.33cm}||m{.25cm}|m{.25cm}|m{.25cm}|m{.25cm}|m{.25cm}||m{.25cm}|m{.25cm}|m{.25cm}|m{.25cm}|m{.25cm}||m{.25cm}|m{.25cm}|m{.25cm}|m{.25cm}|m{.25cm}||m{.25cm}|m{.25cm}|m{.25cm}|m{.25cm}|m{.25cm}||>{\centering\arraybackslash}m{3cm}|}
    \hline & \multicolumn{5}{c||}{FlashAttention} & \multicolumn{5}{c||}{Hydragen} & \multicolumn{5}{c||}{vLLM (No Tokenization)} & \multicolumn{5}{c||}{vLLM} & \multicolumn{1}{c|}{Upper Bound (No Attention)}  \\
    \hline Batch & \multicolumn{5}{c||}{Prefix length} & \multicolumn{5}{c||}{Prefix length} & \multicolumn{5}{c||}{Prefix length} & \multicolumn{5}{c||}{Prefix length} & \multicolumn{1}{c|}{Prefix length}  \\
    \cline{2-22} Size & \vspace{0.02cm} $1$K\vspace{0.02cm} & \vspace{0.02cm} $2$K\vspace{0.02cm} & \vspace{0.02cm} $4$K\vspace{0.02cm} & \vspace{0.02cm} $8$K\vspace{0.02cm} & \vspace{0.02cm} $16$K\vspace{0.02cm} & \vspace{0.02cm} $1$K\vspace{0.02cm} & \vspace{0.02cm} $2$K\vspace{0.02cm} & \vspace{0.02cm} $4$K\vspace{0.02cm} & \vspace{0.02cm} $8$K\vspace{0.02cm} & \vspace{0.02cm} $16$K\vspace{0.02cm} & \vspace{0.02cm} $1$K\vspace{0.02cm} & \vspace{0.02cm} $2$K\vspace{0.02cm} & \vspace{0.02cm} $4$K\vspace{0.02cm} & \vspace{0.02cm} $8$K\vspace{0.02cm} & \vspace{0.02cm} $16$K\vspace{0.02cm} & \vspace{0.02cm} $1$K\vspace{0.02cm} & \vspace{0.02cm} $2$K\vspace{0.02cm} & \vspace{0.02cm} $4$K\vspace{0.02cm} & \vspace{0.02cm} $8$K\vspace{0.02cm} & \vspace{0.02cm} $16$K\vspace{0.02cm}& All \\
    \hline 32 & 1.4\newline$\pm$\newline0.0 & 1.4\newline$\pm$\newline0.0 & 1.2\newline$\pm$\newline0.0 & 1.0\newline$\pm$\newline0.0 & 0.8\newline$\pm$\newline0.0 & 1.4\newline$\pm$\newline0.0 & 1.4\newline$\pm$\newline0.0 & 1.4\newline$\pm$\newline0.0 & 1.4\newline$\pm$\newline0.0 & 1.4\newline$\pm$\newline0.0 & 1.5\newline$\pm$\newline0.0 & 1.4\newline$\pm$\newline0.0 & 1.2\newline$\pm$\newline0.0 & 0.5\newline$\pm$\newline0.0 & 0.3\newline$\pm$\newline0.0 & 1.5\newline$\pm$\newline0.0 & 1.3\newline$\pm$\newline0.0 & 1.1\newline$\pm$\newline0.0 & 0.5\newline$\pm$\newline0.0 & 0.3\newline$\pm$\newline0.0 & $1.6\pm0.0$\\
 \hline64 & 2.5\newline$\pm$\newline0.0 & 2.3\newline$\pm$\newline0.1 & 2.1\newline$\pm$\newline0.0 & 1.8\newline$\pm$\newline0.0 & 1.3\newline$\pm$\newline0.0 & 2.6\newline$\pm$\newline0.0 & 2.6\newline$\pm$\newline0.0 & 2.5\newline$\pm$\newline0.0 & 2.5\newline$\pm$\newline0.0 & 2.5\newline$\pm$\newline0.0 & 2.6\newline$\pm$\newline0.0 & 2.3\newline$\pm$\newline0.0 & 1.9\newline$\pm$\newline0.0 & 0.7\newline$\pm$\newline0.0 & 0.4\newline$\pm$\newline0.0 & 2.4\newline$\pm$\newline0.0 & 2.1\newline$\pm$\newline0.1 & 1.6\newline$\pm$\newline0.0 & 0.6\newline$\pm$\newline0.0 & 0.4\newline$\pm$\newline0.0 & $2.9\pm0.0$\\
 \hline128 & 3.8\newline$\pm$\newline0.0 & 3.4\newline$\pm$\newline0.0 & 2.8\newline$\pm$\newline0.0 & 2.1\newline$\pm$\newline0.0 & X & 4.2\newline$\pm$\newline0.0 & 4.1\newline$\pm$\newline0.0 & 4.1\newline$\pm$\newline0.0 & 4.0\newline$\pm$\newline0.0 & 3.9\newline$\pm$\newline0.0 & 3.8 & 3.0 & 2.3 & 0.8 & 0.4 & 3.4 & 2.7 & 2.0 & 0.7 & 0.4 & $4.4\pm0.3$\\
 \hline256 & 6.0\newline$\pm$\newline0.0 & 5.3\newline$\pm$\newline0.0 & 4.4\newline$\pm$\newline0.0 & X & X & 6.6\newline$\pm$\newline0.0 & 6.6\newline$\pm$\newline0.0 & 6.5\newline$\pm$\newline0.0 & 6.3\newline$\pm$\newline0.0 & 5.9\newline$\pm$\newline0.0 & 5.1 & 3.9 & 2.8 & 0.8 & 0.4 & 4.4 & 3.3 & 2.4 & 0.8 & 0.4 & $7.2\pm0.2$\\
 \hline512 & 7.0\newline$\pm$\newline0.0 & 6.0\newline$\pm$\newline0.0 & X & X & X & 8.2\newline$\pm$\newline0.0 & 8.1\newline$\pm$\newline0.0 & 8.0\newline$\pm$\newline0.0 & 7.8\newline$\pm$\newline0.0 & 7.3\newline$\pm$\newline0.0 & 4.2 & 2.7 & 1.5 & 0.8 & 0.4 & 3.6 & 2.4 & 1.4 & 0.8 & 0.4 & $8.8\pm0.1$\\
 \hline1024 & X & X & X & X & X & 9.4\newline$\pm$\newline0.0 & 9.2\newline$\pm$\newline0.0 & 9.0\newline$\pm$\newline0.0 & 8.5\newline$\pm$\newline0.0 & 7.6\newline$\pm$\newline0.0 & 4.3 & 2.8 & 1.6 & 0.8 & 0.4 & 3.7 & 2.5 & 1.4 & 0.8 & 0.4 & $9.9\pm0.2$\\
 \hline2048 & X & X & X & X & X & 10.4\newline$\pm$\newline0.0 & 10.3\newline$\pm$\newline0.0 & 10.0\newline$\pm$\newline0.0 & 9.4\newline$\pm$\newline0.0 & 8.5\newline$\pm$\newline0.0 & 4.3 & 2.7 & 1.5 & 0.8 & 0.4 & 3.7 & 2.4 & 1.4 & 0.8 & 0.4 & $11.0\pm0.0$\\
 \hline4096 & X & X & X & X & X & 11.1\newline$\pm$\newline0.0 & 11.0\newline$\pm$\newline0.0 & 10.7\newline$\pm$\newline0.0 & 10.2\newline$\pm$\newline0.0 & 9.4\newline$\pm$\newline0.0 & 4.0 & 2.6 & 1.4 & 0.8 & 0.4 & 3.5 & 2.3 & 1.3 & 0.7 & 0.4 & $11.6\pm0.0$\\ \hline
    \end{tabular}}
    \caption{End-to-end decoding throughput (thousands of tokens per second) with CodeLlama-34B on 8xA100 40 GB GPUs when generating 128 tokens. An x indicates the model does not have the required memory to run.}
    \label{tab:fsynth-34b}
    \end{table}

    \begin{table}[H]\centering\tiny\noindent\makebox[\textwidth]{\begin{tabular}{|m{0.33cm}||m{.25cm}|m{.25cm}|m{.25cm}|m{.25cm}|m{.25cm}||m{.25cm}|m{.25cm}|m{.25cm}|m{.25cm}|m{.25cm}||m{.25cm}|m{.25cm}|m{.25cm}|m{.25cm}|m{.25cm}||m{.25cm}|m{.25cm}|m{.25cm}|m{.25cm}|m{.25cm}||>{\centering\arraybackslash}m{3cm}|}
    \hline & \multicolumn{5}{c||}{FlashAttention} & \multicolumn{5}{c||}{Hydragen} & \multicolumn{5}{c||}{vLLM (No Tokenization)} & \multicolumn{5}{c||}{vLLM} & \multicolumn{1}{c|}{Upper Bound (No Attention)}  \\
    \hline Batch & \multicolumn{5}{c||}{Prefix length} & \multicolumn{5}{c||}{Prefix length} & \multicolumn{5}{c||}{Prefix length} & \multicolumn{5}{c||}{Prefix length} & \multicolumn{1}{c|}{Prefix length}  \\
    \cline{2-22} Size & \vspace{0.02cm} $1$K\vspace{0.02cm} & \vspace{0.02cm} $2$K\vspace{0.02cm} & \vspace{0.02cm} $4$K\vspace{0.02cm} & \vspace{0.02cm} $8$K\vspace{0.02cm} & \vspace{0.02cm} $16$K\vspace{0.02cm} & \vspace{0.02cm} $1$K\vspace{0.02cm} & \vspace{0.02cm} $2$K\vspace{0.02cm} & \vspace{0.02cm} $4$K\vspace{0.02cm} & \vspace{0.02cm} $8$K\vspace{0.02cm} & \vspace{0.02cm} $16$K\vspace{0.02cm} & \vspace{0.02cm} $1$K\vspace{0.02cm} & \vspace{0.02cm} $2$K\vspace{0.02cm} & \vspace{0.02cm} $4$K\vspace{0.02cm} & \vspace{0.02cm} $8$K\vspace{0.02cm} & \vspace{0.02cm} $16$K\vspace{0.02cm} & \vspace{0.02cm} $1$K\vspace{0.02cm} & \vspace{0.02cm} $2$K\vspace{0.02cm} & \vspace{0.02cm} $4$K\vspace{0.02cm} & \vspace{0.02cm} $8$K\vspace{0.02cm} & \vspace{0.02cm} $16$K\vspace{0.02cm}& All \\
    \hline 32 & 1.4\newline$\pm$\newline0.0 & 1.3\newline$\pm$\newline0.0 & 1.2\newline$\pm$\newline0.0 & 1.1\newline$\pm$\newline0.0 & 0.8\newline$\pm$\newline0.0 & 1.4\newline$\pm$\newline0.0 & 1.4\newline$\pm$\newline0.0 & 1.4\newline$\pm$\newline0.0 & 1.4\newline$\pm$\newline0.0 & 1.3\newline$\pm$\newline0.0 & 1.5\newline$\pm$\newline0.0 & 1.4\newline$\pm$\newline0.0 & 1.2\newline$\pm$\newline0.0 & 0.5\newline$\pm$\newline0.0 & 0.3\newline$\pm$\newline0.0 & 1.5\newline$\pm$\newline0.0 & 1.3\newline$\pm$\newline0.1 & 1.1\newline$\pm$\newline0.0 & 0.5\newline$\pm$\newline0.0 & 0.3\newline$\pm$\newline0.0 & $1.5\pm0.1$\\
 \hline64 & 2.5\newline$\pm$\newline0.0 & 2.4\newline$\pm$\newline0.0 & 2.1\newline$\pm$\newline0.0 & 1.8\newline$\pm$\newline0.0 & 1.3\newline$\pm$\newline0.0 & 2.5\newline$\pm$\newline0.0 & 2.5\newline$\pm$\newline0.0 & 2.5\newline$\pm$\newline0.0 & 2.5\newline$\pm$\newline0.0 & 2.4\newline$\pm$\newline0.0 & 2.6\newline$\pm$\newline0.0 & 2.3\newline$\pm$\newline0.0 & 1.8\newline$\pm$\newline0.0 & 0.7\newline$\pm$\newline0.0 & 0.4\newline$\pm$\newline0.0 & 2.3\newline$\pm$\newline0.1 & 2.0\newline$\pm$\newline0.0 & 1.6\newline$\pm$\newline0.0 & 0.6\newline$\pm$\newline0.0 & 0.4\newline$\pm$\newline0.0 & $2.8\pm0.1$\\
 \hline128 & 3.8\newline$\pm$\newline0.0 & 3.4\newline$\pm$\newline0.0 & 2.8\newline$\pm$\newline0.0 & 2.1\newline$\pm$\newline0.0 & X & 4.1\newline$\pm$\newline0.0 & 4.1\newline$\pm$\newline0.0 & 4.0\newline$\pm$\newline0.0 & 4.0\newline$\pm$\newline0.0 & 3.8\newline$\pm$\newline0.0 & 3.7 & 3.0 & 2.2 & 0.7 & 0.4 & 3.2 & 2.6 & 2.0 & 0.7 & 0.4 & $4.5\pm0.1$\\
 \hline256 & 5.8\newline$\pm$\newline0.0 & 5.3\newline$\pm$\newline0.0 & 4.3\newline$\pm$\newline0.0 & X & X & 6.5\newline$\pm$\newline0.0 & 6.5\newline$\pm$\newline0.0 & 6.4\newline$\pm$\newline0.0 & 6.2\newline$\pm$\newline0.0 & 5.8\newline$\pm$\newline0.0 & 5.0 & 3.9 & 2.7 & 0.8 & 0.4 & 4.2 & 3.3 & 2.3 & 0.7 & 0.4 & $7.1\pm0.2$\\
 \hline512 & 6.8\newline$\pm$\newline0.0 & 5.9\newline$\pm$\newline0.0 & X & X & X & 8.0\newline$\pm$\newline0.0 & 8.0\newline$\pm$\newline0.0 & 7.9\newline$\pm$\newline0.0 & 7.6\newline$\pm$\newline0.0 & 7.2\newline$\pm$\newline0.0 & 3.9 & 2.6 & 1.5 & 0.8 & 0.4 & 3.5 & 2.3 & 1.4 & 0.7 & 0.4 & $8.8\pm0.1$\\
 \hline1024 & X & X & X & X & X & 9.2\newline$\pm$\newline0.0 & 9.1\newline$\pm$\newline0.0 & 8.8\newline$\pm$\newline0.0 & 8.3\newline$\pm$\newline0.0 & 7.5\newline$\pm$\newline0.0 & 3.9 & 2.6 & 1.4 & 0.8 & 0.4 & 3.6 & 2.4 & 1.4 & 0.7 & 0.4 & $9.9\pm0.0$\\
 \hline2048 & X & X & X & X & X & 10.3\newline$\pm$\newline0.0 & 10.1\newline$\pm$\newline0.0 & 9.8\newline$\pm$\newline0.0 & 9.3\newline$\pm$\newline0.0 & 8.4\newline$\pm$\newline0.0 & 4.0 & 2.6 & 1.5 & 0.8 & 0.4 & 3.6 & 2.4 & 1.4 & 0.7 & 0.4 & $11.0\pm0.0$\\ \hline
    \end{tabular}}
    \caption{End-to-end decoding throughput (thousands of tokens per second) with CodeLlama-34B on 8xA100 40 GB GPUs when generating 256 tokens. An x indicates the model does not have the required memory to run.}
    \label{tab:fsynth-34b-c256}
    \end{table}

\subsection{Microbenchmarks}
\label{app:micro}

\begin{figure}
    \centering
  \includegraphics[width=0.45\linewidth]{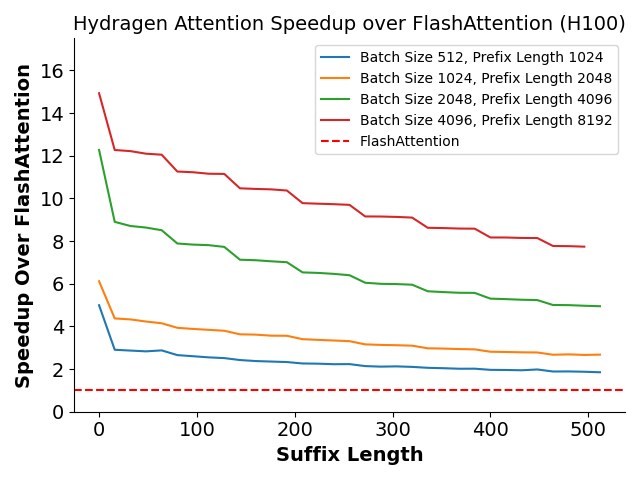}
  \includegraphics[width=0.45\linewidth]{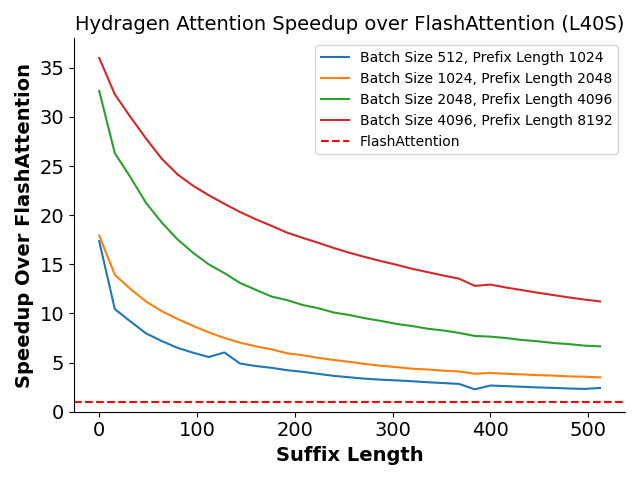}
    \caption{Speedup of Hydragen attention over FlashAttention for various batch sizes, shared prefix lengths and suffix lengths on an H100 (left) and an L40S (right) GPU.}
    \label{fig:micro_moregpu}
\label{fig:test}
\end{figure}

We repeat the A100 microbenchmark experiment from Section~\ref{sec:micro} on H100 and L40S GPUs, reporting our results in Figure~\ref{fig:micro_moregpu}. The L40S has the highest ratio of FLOPs to memory bandwidth of the three GPUs and therefore derives the most benefit from Hydragen's elimination of memory bottlenecks. While the compute-to-bandwidth ratio is higher on an H100 than on an A100, we measure similar speedups on both cards. This stems from the fact that the \verb|flash-attn| package that we use is not currently optimized for Hopper GPUs, and therefore achieves a lower device utilization on an H100 vs an A100. 

\section{Experiment Details}

\subsection{End-to-End Benchmarks}
\label{app:e2e-details}

Our end-to-end benchmarks only measure decoding throughput and exclude the time required to compute the prefill. We measure ``decode-only'' time by initially benchmarking the time required to generate one token from a given prompt and subtracting that value from the time it takes to generate the desired number of tokens. This subtraction is particularly important in order to fairly evaluate vLLM baselines, since it appears that vLLM redundantly detokenizes the prompt for every sequence in the batch at the beginning of inference (this can take minutes for large batch sizes and sequence lengths). For our ``vLLM no detokenization'' baseline, we disable incremental detokenization in vLLM by commenting out \href{https://github.com/vllm-project/vllm/blob/2e0b6e775756345aa1d39f772c186e00f8c29e92/vllm/engine/llm_engine.py#L468}{this line}.

For all FlashAttention and No Attention datapoints, we run 10 warmup iterations and use the following 10 iterations to compute throughput. For Hydragen datapoints, we run 10 warmup and 10 timing iterations when the batch size is less than 256, and for larger batch sizes we use three warmup and three timing iterations. We observe that shorter-running Hydragen benchmarks (those with smaller batch sizes, sequence lengths, model sizes, or completion lengths) can occasionally produce longer outlier times. This seems to be related not to decoding time itself, but to variations in prefilling time before decoding. For vLLM baselines (both with and without incremental detokenization), we use three warmup and timing iterations for all batch sizes below 128, as well as for all datapoints that are used in Figures~\ref{fig:synth-xbs} and \ref{fig:synth-xp}. The longest-running vLLM runs can take many minutes to complete a single iteration, so for baselines above a batch size of 128 that only appear in the supplementary tables of Appendix~\ref{app:e2e}, we use one warmup and one timing iteration.

\subsection{Microbenchmarks}
\label{app:micro-details}

In each microbenchmark, we run 1000 iterations of warmup before reporting the mean running time across 1000 trials. Between iterations, we flush the GPU L2 cache by writing to a 128MiB tensor. We use CUDA graphs when benchmarking in order to reduce CPU overhead, which can be important since some benchmarks can complete a single iteration in tens of microseconds.

\subsection{Long document retrieval}
\label{app:needles-details}

To demonstrate the throughput benefits of using Hydragen to answer questions about a long document, we construct a document (with 19974 tokens) that contains arbitrary facts from which question/answer pairs can be easily generated. 

\textbf{Prefix and Suffix Content:} The content of the document is a subset of \textit{War and Peace} \cite{Tolstoy1869}, modified to include procedurally generated facts of the form ``The dog named \{name\} has fur that is \{color\}''. The questions are of the form ``What color is the fur of the dog named {name}?'', where the answer is \{color\}. We construct 261 questions (256 testable questions plus five for the few-shot examples) and interleave these throughout sentences of the document. When benchmarking with a greater number of questions than 256, we duplicate questions when querying the model - this is instead of adding more questions to the document in order to constrain total document length.

\textbf{Model and Accelerator Choice:} We choose the Yi-6B-200k model because it is small enough to fit a large KV cache in memory (important when running baselines that redundantly store the document) while also supporting a long enough context to process our document. We distribute the model across four A100-40GB GPUs in order to maximize possible KV cache size (the model only has four key/value attention heads, preventing us from easily using tensor parallelism across more GPUs). Our reported measurements use the mean of five timing runs after ten warmup iterations.

\subsection{Hierarchical Sharing in Competitive Programming}
\label{app:hierarchy-details}

The dataset of 120 problems that we use for this benchmark comes from the introductory difficulty split of APPS. We filter out problems that include starter code. We use two few-shot examples (2400 tokens long) that come from the training split of APPS, while all of the eval examples come from the test split. We sample 512 tokens for every completion. We run this experiment using CodeLlama-7b on eight A100-40GB GPUs. We measure the total time to run inference on all 120 questions, excluding tokenization and detokenization time.

\end{document}